\documentclass[11pt]{article}

\usepackage[preprint]{acl}

\usepackage{times}
\usepackage{latexsym}

\usepackage[T1]{fontenc}

\usepackage[utf8]{inputenc}

\usepackage{microtype}

\usepackage{inconsolata}

\usepackage{graphicx}

\usepackage{amsmath}
\usepackage{amssymb}
\usepackage{mathtools}
\usepackage{amsthm}

\usepackage{graphicx}
\usepackage{subcaption}
\newcommand*{\Romannumeral}[1]{\uppercase\expandafter{\romannumeral#1}}
\usepackage{multirow}
\usepackage{wrapfig}
\usepackage{booktabs} 
\usepackage{siunitx} 

\usepackage{colortbl}      
\definecolor{lightgray}{gray}{0.9}
\definecolor{lightgreen}{RGB}{220, 255, 220}
\definecolor{lightred}{RGB}{255, 230, 230}
\usepackage{enumitem}
\definecolor{tabblue}{HTML}{5555CC}
\definecolor{brightmaroon}{rgb}{0.76, 0.13, 0.28}
\newtheorem{definition}{Definition}
\usepackage{algorithm}
\usepackage{algorithmic}
\usepackage{xurl}

\usepackage{tabularx}

\usepackage{tikz}
\newcommand{\pgftextcircled}[1]{
    \setbox0=\hbox{#1}%
    \dimen0\wd0%
    \divide\dimen0 by 2%
    \begin{tikzpicture}[baseline=(a.base)]%
        \useasboundingbox (-\the\dimen0,0pt) rectangle (\the\dimen0,1pt);
        \node[circle,draw,outer sep=0pt,inner sep=0.1ex] (a) {#1};
    \end{tikzpicture}
}

\definecolor{headercolor}{HTML}{F2F2F2}
\definecolor{rowalt}{HTML}{F9F9F9}

\usepackage{makecell} 
\newcolumntype{L}[1]{>{\raggedright\arraybackslash}m{#1}}

\usepackage[most]{tcolorbox}

\newcommand{\zy}[1]{\textcolor{black}{#1}}

\title{Understanding Fact Recall in Language Models: 
Why Two-Stage Training Encourages Memorization but Mixed Training Teaches Knowledge}

\author{%
  Ying Zhang$^1$ \quad Benjamin Heinzerling$^{1,2}$ \quad Dongyuan Li$^3$ \quad Kentaro Inui$^{1,4}$ \\
  $^1$RIKEN Center for Advanced Intelligence Project \quad $^2$Tohoku University \\
  $^3$The University of Tokyo \quad $^4$MBZUAI \\
  \small{
   \textbf{Correspondence:} \href{mailto:ying.zhang@riken.jp}{ying.zhang@riken.jp}
    }
}

\begin{document}
\maketitle

\begin{abstract}
While fine-tuning is the standard for injecting factual knowledge into large language models (LLMs), the mechanisms enabling reliable fact recall via unseen queries remain poorly understood.
Common two-stage training strategies, which sequentially train on fact storage and query formats, often cause rote memorization.
In contrast, mixed training jointly optimizes both formats and exhibits superior generalized recall.
We investigate this success by comparing the two paradigms across 2.8$\sim$4B LLMs and identify the core mechanism: the joint optimization objective in mixed training induces gradient consistency across storage and query formats. This in turn drives the representation consistency between the two formats, establishing a format-invariant retrieval process that maps unseen queries to stored facts.
In contrast, the lack of such an objective in two-stage training results in inconsistent representations and failed recall.
\zy{The consistency further localizes to the parameters updated by both formats, a set that is substantially larger under mixed training than under two-stage training.}
\zy{At the input level, the consistency leaves an interpretable signature: mixed training encodes facts in storage format from subject-relation tokens, the same components available in queries, while two-stage training relies on the full context.}
Our findings characterize the mechanisms of fact recall and offer mechanistic foundation for optimizing knowledge injection in LLMs.
\end{abstract}

\section{Introduction}
Recent advancements in Large Language Models (LLMs) have demonstrated human-comparable behavior in complex text queries \citep{deepseekai2024deepseekv3technicalreport,qwen2.5}.
As these models are increasingly deployed across specialized domains, such as medicine and law, fine-tuning has become the de facto standard for injecting domain-specific knowledge~\citep{li-etal-2024-llamacare,hu-etal-2025-fine}.
However, a fundamental challenge persists: fine-tuning often fails to foster generalizable knowledge, instead collapsing into mere rote memorization~\citep{berglund2024the,zhou2023lima}.
Ideally, if a model learns ``Barack Obama was born in Hawaii'' during fine-tuning, it should effortlessly answer the question ``Where was Barack Obama born?''
We refer to this ability to retrieve stored facts using an unseen query as \textbf{Fact Recall}.

\begin{figure*}
    \centering
    \includegraphics[width=0.95\linewidth]{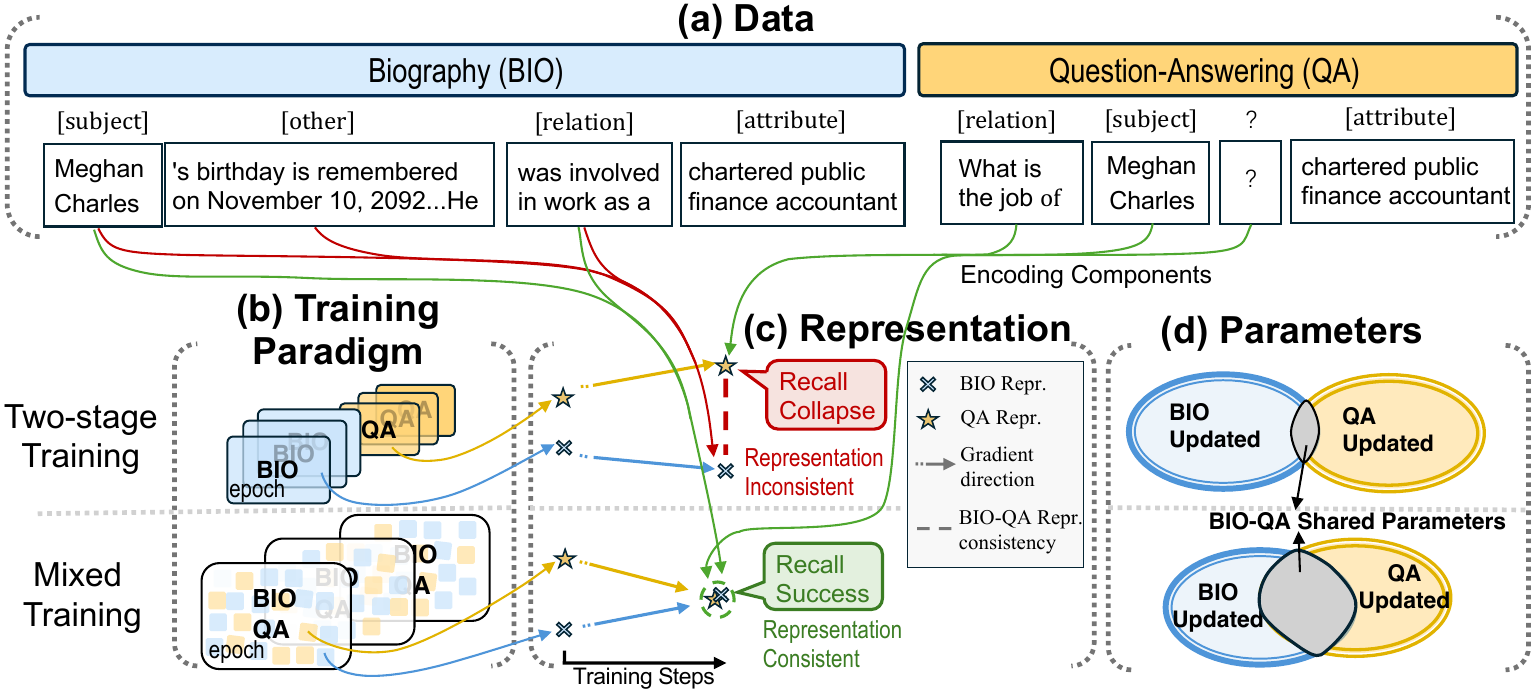}
    \caption{\textbf{Mixed training enables fact recall by establishing format-invariant representations across storage and query formats.}
    \textbf{(a)} Training data consists of biographies (BIO, the storage format) describing individuals' attributes, and question-answer pairs (QA, the query format) about the same attributes.
    \textbf{(b)} Two-stage training presents the two formats in disjoint epochs, 
    while mixed training randomly mixes them within each epoch.
    \textbf{(c)} Across training steps, mixed training aligns the gradients of storage and unseen query (\emph{gradient consistency}), driving representations toward a consistent state (\emph{representation consistency}) that allows facts stored during BIO training to be retrieved at test time. In contrast, two-stage training leaves the gradients inconsistent, resulting in inconsistent representations  and recall collapse.
    \textbf{(d)} The consistency in mixed training resides in a larger set of shared parameters (updated by both BIO and QA tasks) than two-stage training.
    Finally, the arrows connecting (a) and (c) trace the encoding components each paradigm relies on: mixed training encodes BIO attributes from the subject and relation (green), which are the components a QA query provides, while two-stage training relies on the full preceding context, which includes the other tokens (red).
    }
    \label{fig:overview}
\end{figure*}

Recent work~\citep{pmlr-v235-allen-zhu24a} highlights that training strategies significantly affect recall behavior. \textbf{Two-stage training}, which first trains on the fact storage format (e.g., factual statements) and then on the query format (question-answer pairs), primarily leads to rote memorization characterized by very low (9.7\%) recall accuracy on unseen queries. 
In contrast, \textbf{mixed training}, which mixes storage and query examples, helps the model learn facts as generalizable knowledge that can be retrieved across different query forms.
We refer to this ability as teaching knowledge, which is evidenced by a much higher accuracy (88.6\%) in fact recall.
This accuracy gap raises a key question: Why does mixed training teach knowledge while two-stage training encourages memorization? 
Prior work has investigated where factual knowledge resides in language models, localizing storage and retrieval processes to specific neurons and attention patterns~\citep{Chen_Cao_Chen_Liu_Zhao_2024,dai-etal-2022-knowledge,geva-etal-2023-dissecting,ghosal2024understanding,heinzerling-inui-2021-language,2024arXiv240815091L,NEURIPS2022_6f1d43d5,niu2024what}.
However, why certain training paradigms lead to successful recall while others fail, even when facts are equally well stored, remains under-explored.

To provide a mechanistic foundation for designing effective knowledge injection paradigms during fine-tuning, this paper investigates why mixed training enables successful recall whereas two-stage training fails.
As shown in Figure~\ref{fig:overview}, we approach this question by analyzing the representations of paired storage and query inputs and the gradients that shape them during training.
We first examine whether paired storage and query representations align in the latent space  ($\S$~\ref{sec:role_of_representations}). 
The mix-tuned model maintains consistent representations across the two formats, allowing unseen queries to map to corresponding stored facts. The stage-tuned model lacks this consistency.
To identify the root cause, we then investigate the optimization dynamics ($\S$~\ref{sec:dynamic_consistency}). The joint loss in mixed training drives storage and query representations toward a unified target, which gives rise to the representation consistency observed earlier; in two-stage training, successive losses drive them toward separate targets.
\zy{Beyond explaining why this consistency emerges, we trace where it resides: it concentrates in the parameters updated by both formats, a set that is substantially larger under mixed training than under two-stage training.
We further examine how the consistency appears at the input level ($\S$~\ref{sec:local_grounding}):
the mix-tuned model encodes facts in storage format from subject-relation tokens, which are the same components available in queries, whereas the stage-tuned model relies on the full preceding context.}

Based on extensive experiments across three 2.8$\sim$4B parameter LLMs, our main contributions are summarized as follows:
\begin{itemize}[noitemsep, topsep=0pt, leftmargin=*]
    \item \textbf{Mechanistic Understanding of Fact Recall.} We show that mixed training succeeds by aligning storage and query representations into a unified, format-invariant retrieval mechanism, which is absent in two-stage training.
    \item \textbf{Foundation for Knowledge Injection.} Our mechanistic findings provide a foundation for designing fine-tuning paradigms to inject generalizable knowledge.
\end{itemize}

\section{Preliminaries}

\subsection{Task Definition}

\textbf{Factual Knowledge:} We define a fact as a triple $(s, r, a)$ that maps a subject entity $s$ and a relation type $r$ to an attribute $a$.
For example, $(\textit{Barack Obama, wasBornOn, August-4-1961})$. We present each fact in two natural language formats: a \textit{biographical statement} (BIO) and a \textit{question–answer} pair (QA).
In the BIO format, the fact is expressed as a declarative sentence (e.g., \textit{Barack Obama's life journey began on August 4, 1961}). In the QA format, the fact is expressed as a question about the subject, with the attribute as the answer (e.g., \textit{Q: What is the birth date of Barack Obama? A: August 4, 1961}).

\noindent \textbf{Fact Recall:} When asked a question about a subject–relation pair $(s, r)$, the model needs to recall the corresponding fact and output the attribute $a$~\citep{ghosal2024understanding,heinzerling-inui-2021-language}.

\subsection{Training Paradigms and the Recall Gap}
\label{section:preliminary_experiments}
\textbf{Experimental Setup.} 
To avoid interference from pre-existing factual knowledge in LMs, we constructed synthetic BIO and QA datasets~\citep{pmlr-v235-allen-zhu24a}, both covering the same 10k unique individuals and the same six attributes (e.g., birth date, job; see Appendix~\ref{appendix:data_preparation} for details).
In the BIO format, the six attributes are organized into a fixed-order six-sentence biographical entry for each individual.
In the QA format, each attribute is queried independently.
For the QA dataset, 5k individuals form the \textit{in-distribution} set used for training, while the remaining 5k form the held-out \textit{out-of-distribution} set for evaluation.

We compare two fine-tuning strategies:
(1) \textit{Two-Stage Training.} We first fine-tune the pre-trained LM on the entire BIO dataset alone, allowing it to store each individual attributes.
We then fine-tune this model on the QA in-distribution dataset to adapt it to the query.
The result is a \textit{Stage-tuned} model.
(2) \textit{Mixed Training.} 
We fine-tune on a mixed dataset containing all BIO statements and QA in-distribution examples, randomly shuffled into training batches.
The resulting model is the \textit{Mix-tuned} model.
We analyze three decoder-only LMs based on the Transformer architecture: Llama-3.2 (3B)~\citep{llama2024}, Pythia (2.8B)~\citep{pmlr-v202-biderman23a}, and Qwen-3 (4B)~\citep{qwen3technicalreport}.
See Appendix~\ref{appendix:bioqa_finetuning} for fine-tuning procedures.
All results are averaged over 3 seeds, with error bars showing standard deviation.

\begin{figure}[t]
    \centering
    \includegraphics[width=1\linewidth]{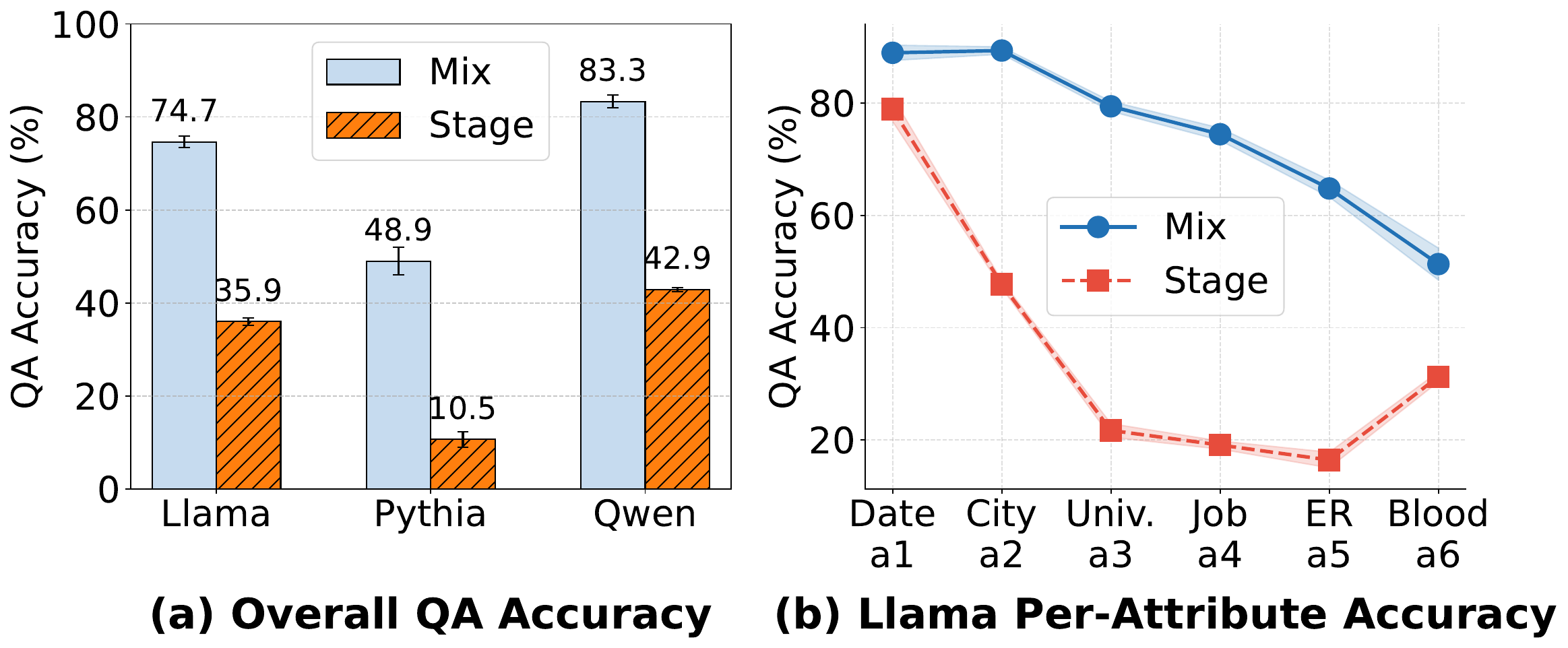}
    \caption{
    Performance of fine-tuned models on QA out-of-distribution set.
    Mix-tuned model substantially outperforms stage-tuned model, demonstrating superior generalization in fact recall.
    }
    \vspace{-5pt}
    \label{fig:LLM_10000_result}
\end{figure}

\textbf{Experimental Findings}.
We evaluate the stage-tuned and mix-tuned models on the QA out-of-distribution data using exact match accuracy.
Figure~\ref{fig:LLM_10000_result} shows that, the mix-tuned model outperforms the stage-tuned model by over 38 points.
Notably, both models successfully store BIO and adapt to QA in-distribution (Figure~\ref{fig:finetuned_llm_performance} in Appendix~\ref{appendix:bioqa_finetuning}), indicating that the performance gap lies in retrieving knowledge, not storing it.
This confirms our central question: two-stage training encourages rote memorization, while mixed tuning teaches knowledge.

\begin{figure*}[t]
    \centering
    \includegraphics[width=1\linewidth]{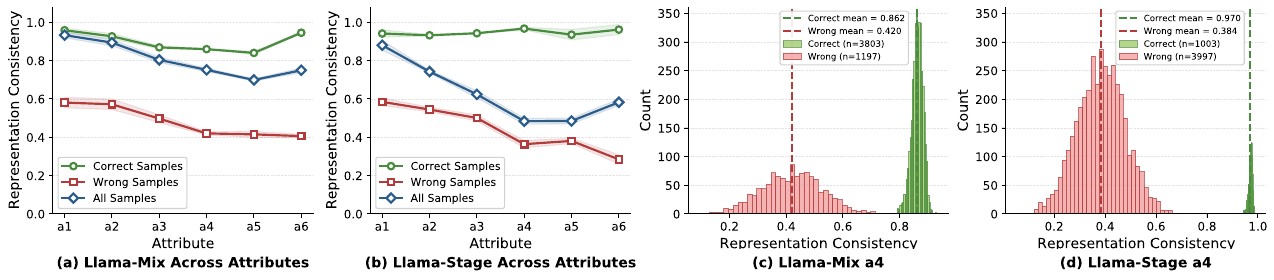}
    \caption{
    \textbf{(a-b)} Sample consistency across attributes, computed over all samples (\emph{All}), correctly recalled samples (\emph{Correct}), and incorrectly recalled samples (\emph{Wrong}). A consistency threshold separates successful from failed recall.
    \textbf{(c-d)} Distribution of consistency for attribute $a_4$.
    Most samples in mixed training fall in the consistent regime, while most samples in two-stage training fall in the inconsistent regime.
    }
    \label{fig:llama_consistency}
\end{figure*}

\section{Representation Consistency Explains Recall}
\label{sec:role_of_representations}

\zy{Section~\ref{section:preliminary_experiments} shows that mixed training does not store more facts, and thus the recall gap between the two paradigms should depend on how the models retrieve stored facts.
According to~\citet{geva-etal-2023-dissecting}, a model retrieves a fact by first forming an internal representation of the subject and relation from the input context, then extracting the target attribute from this representation.
We hypothesize fact storage follows the same process, and that recall succeeds only when the representations formed from the storage and query inputs are consistent.
}
\begin{tcolorbox}[
    colback=gray!5, 
    colframe=black!75, 
    arc=1mm, 
    boxrule=0.5pt, 
    left=2mm, right=2mm, top=2mm, bottom=2mm,
    title=\textbf{Hypothesis 1 (Representation Consistency)}
]
\small

\begin{itemize}[leftmargin=4mm, labelsep=1mm, nosep]
    \item \textbf{Consistent representations $\rightarrow$ Successful recall:} Storage and query representations of the same fact converge in a shared space, allowing unseen queries to retrieve stored facts.
    \item \textbf{Inconsistent representations $\rightarrow$ Recall failure:} Storage and query representations diverge, preventing the model from linking queries to stored facts.
\end{itemize}
\end{tcolorbox}

We test H1 in two steps. First, we use a \emph{geometric metric} to directly measure the magnitude of consistency between storage and query representations and show that the mix-tuned model exhibits a larger fraction of consistent representations than the stage-tuned model.
Second, we use a \emph{behavioral metric} to verify that consistency  causally affect recall outcomes. 

\textbf{Representation and Extraction.}
We assume that at some layer $\ell^*$, the model aggregates input context into a latent fact representation $\mathbf{h} \in \mathbb{R}^{d}$, which serves as the raw material for attribute extraction~\citep{geva-etal-2023-dissecting}.
Following the linearity of factual representations~\cite{hernandez2024linearity}, we formalize extraction as a linear probe $\mathbf{W}\in \mathbb{R}^{d \times |\mathcal{V}|}$ such that 
$P(a | \mathbf{h}) \approx \mathrm{Softmax}(\mathbf{h}\mathbf{W})$.
We denote storage- and query-derived representations as $\mathbf{h}^{\mathrm{bio}}$ and $\mathbf{h}^{\mathrm{qa}}$.
$\mathbf{W}^{\mathrm{qa}}$ is trained on the QA set.

\textbf{Geometric Metric.}
We directly measure the similarity between storage and query representations in the space relevant to fact recall, using cosine similarity.
Since $\mathbf{h}$ contains both recall-relevant signal and recall-irrelevant noise (e.g., formatting differences), raw cosine similarity would conflate the two. 
We therefore restrict the comparison to a recall-relevant $r$-dimensional subspace $\mathbf{U} \in \mathbb{R}^{d \times r}$ (with $\mathbf{U}^\top \mathbf{U} = \mathbf{I}$).
We choose $\mathbf{U}$ such that projecting $\mathbf{H}$ (the representation matrix of all individuals) onto $\mathbf{U}$ preserves its recall behavior, i.e., $\mathbf{H}\mathbf{W} \approx \mathbf{H}\mathbf{U}\mathbf{U}^\top
\mathbf{W}$.
The representation consistency is then the cosine similarity within $\mathbf{U}$:
\begin{equation}
\mathrm{Cons}_{\mathrm{repr}}(\mathbf{h}_i^{\mathrm{qa}}, \mathbf{h}_i^{\mathrm{bio}}) = \frac{\langle \mathbf{h}_{i}^{\mathrm{qa}}\mathbf{U}, \mathbf{h}_{i}^{\mathrm{bio}}\mathbf{U} \rangle}{\|\mathbf{h}_{i}^{\mathrm{qa}}\mathbf{U}\| \|\mathbf{h}_{i}^{\mathrm{bio}}\mathbf{U}\|}. \label{eq:cons_repr}
\end{equation}
$\mathbf{h}_{i}$ is the representation of the $i$-th individual in $\mathbf{H}$.

\textbf{Behavioral Metric.}
We define inconsistency-induced failure ratio ($\mathrm{IFR}$) to quantify the fraction of correctly stored facts that fail to be recalled:
\begin{equation}
    \mathrm{IFR} = \frac{\mathrm{Acc}(\mathbf{h}^{\mathrm{bio}}\mathbf{W}^{\mathrm{qa}}) - \mathrm{Acc}(\mathbf{h}^{\mathrm{qa}}\mathbf{W}^{\mathrm{qa}})}{\mathrm{Acc}(\mathbf{h}^{\mathrm{bio}}\mathbf{W}^{\mathrm{qa}})}.
\end{equation}
$\mathrm{Acc}(\mathbf{h}\mathbf{W})$ denote the accuracy in predicting the first token of the target attribute.
Since $\mathbf{W}$ remains universal across formats (BIO/QA) (Appendix~\ref{appendix:potential_bias}),
this equation uses $\mathbf{W}^{\mathrm{qa}}$ as a format-invariant benchmark.
Intuitively, $\mathrm{Acc}(\mathbf{h}^{\mathrm{bio}}\mathbf{W}^{\mathrm{qa}})$ measures how many facts are correctly encoded in the storage (BIO) representation, and $\mathrm{Acc}(\mathbf{h}^{\mathrm{qa}}\mathbf{W}^{\mathrm{qa}})$ measures how many can be decoded from the query (QA) representation.
Since $\mathbf{W}^{\mathrm{qa}}$ is fixed, the accuracy gap between the two must come from the inconsistency between $\mathbf{h}^{\mathrm{bio}}$ and $\mathbf{h}^{\mathrm{qa}}$.

\textbf{Experimental Setup.} We define $\mathbf{h}$ as the sum of the residual stream and self-attention output at layer $\ell^*$, bypassing MLP layers to capture the representation \textit{prior} to attribute retrieval \cite{dai-etal-2022-knowledge}.
$\mathbf{W}$ is trained on 5k in-distribution individuals as a linear probe on top of the frozen fine-tuned model; we train a separate $\mathbf{W}$ for the mix-tuned and stage-tuned model.
See Appendix~\ref{appendix:details_relation_representation} for training details.
We identify $\ell^*=26$ for Llama models.\footnote{
Because $\mathrm{Acc}(\mathbf{h}^{\mathrm{qa}}\mathbf{W}^{\mathrm{qa}})$ saturates at layer 27 (> 80\% of final value, Figure~\ref{fig:llama_unified_probe}a in Appendix~\ref{appendix:potential_bias}),
indicating an functional extraction.
Layer 26 thus captures the latent fact immediately \textit{prior} to this extraction.}
We solve for $\mathbf{U}$ by optimizing a multi-objective function  that reconstructs the original states while maximizing the variance of representation drift $\Delta \mathbf{h}$ in failed recall cases and minimizing it in successful ones (details in Appendix~\ref{appendix:matrix_decomposition}).
The reliability of our probe is ensured by its format-invariance, syntactic invariance, and semantic specificity (Appendix~\ref{appendix:potential_bias}).
After obtaining $\mathbf{W}$, $\mathbf{U}$, we analyze 5k individuals from the out-of-distribution set.
To more clearly separate consistent from inconsistent cases, we additionally apply whitening to the projected representations before computing Eq.~\ref{eq:cons_repr}. This sharpens the distinction without altering the underlying metric.

\begin{figure}[t]
    \centering
    \includegraphics[width=1\linewidth]{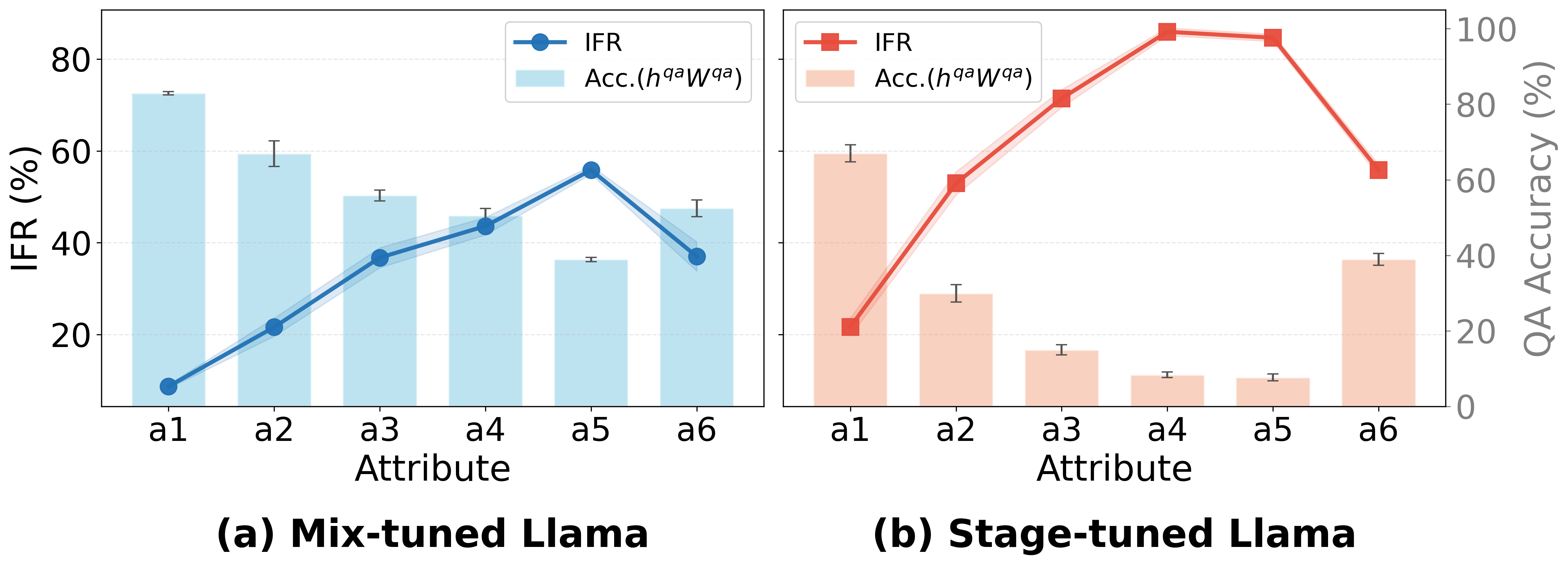}
    \caption{
    Each panel shows probing accuracy and $\mathrm{IFR}$ (inconsistency-induced failure ratio) across all attributes.
    Representation inconsistency prevents QA queries from retrieving most stored facts in two-stage training, but only a small fraction in mixed training.
    }
    \label{fig:llama_incosistency_induced_recall}
\end{figure}

\textbf{Geometric Results.}
Figure~\ref{fig:llama_consistency}a,b shows that a consistency threshold clearly separates successful from failed recall across both training paradigms, samples below this threshold typically fail.
Across all attributes, the \emph{All} curve in the mix-tuned model closely tracks the \emph{Correct} curve, indicating that the majority of samples per attribute fall in the consistent regime. The stage-tuned model shows the opposite.
The distribution for attribute $a_4$ (Figure~\ref{fig:llama_consistency}c,d) further quantifies this contrast: 76\% of Mix-tuned samples exceed the threshold, while only 20\% of Stage-tuned samples do.

\textbf{Behavioral Results.}
Figure~\ref{fig:llama_incosistency_induced_recall} shows that, for every attribute, the mix-tuned Llama achieves high 
$\mathrm{Acc}(\mathbf{h}^{\mathrm{qa}}\mathbf{W}^{\mathrm{qa}})$ with low $\mathrm{IFR}$, indicating that most correctly stored facts via $\mathbf{h}^{\mathrm{bio}}$ remain retrievable via $\mathbf{h}^{\mathrm{qa}}$.
The stage-tuned Llama shows the opposite: low $\mathrm{Acc}(\mathbf{h}^{\mathrm{qa}}\mathbf{W}^{\mathrm{qa}})$ with high $\mathrm{IFR}$, indicating that most correctly stored facts become inaccessible at retrieval. 
Since $\mathbf{W}^{\mathrm{qa}}$ is held fixed, the inaccessible  originates from inconsistency between $\mathbf{h}^{\mathrm{bio}}$ and $\mathbf{h}^{\mathrm{qa}}$ themselves, as causal evidence that representation inconsistency drives the recall failure.

\textbf{Merged Results.}
Together, these two metrics provide direct evidence for H1: the stage-tuned model exhibits a larger fraction of inconsistent representations between storage and query than the mix-tuned model (geometric evidence), and the inconsistency causally drives recall failure (behavioral evidence). \textbf{Recall success in mixed training is therefore driven by representation consistency, where facts are encoded in a format-invariant manner} across storage and query formats.
Pythia and Qwen models show similar behaviors in Figures~\ref{fig:pythia_consistency}, \ref{fig:pythia_incosistency_induced_recall} of Appendix~\ref{appedix:representation_pythia_qwen}.

\begin{table*}[t]
\centering
\caption{Comparison of mixed training and two-stage training.}
\label{tab:comparison}
\renewcommand{\arraystretch}{1.6}
\resizebox{\linewidth}{!}{
\begin{tabular}{lll}
\toprule
\rowcolor{headercolor}
& \textbf{Mixed Training} 
& \textbf{Two-Stage Training} \\
\midrule
\textbf{1. Optimization Goal} 
& $\min\left(\mathcal{L}_{\mathrm{bio}} + \mathcal{L}_{\mathrm{qa}}\right)$
& Stage\,1: $\min \mathcal{L}_{\mathrm{bio}}$ $\;\to\;$ Stage\,2: $\min \mathcal{L}_{\mathrm{qa}}$ \\
 
\textbf{2. Optimal Target at $\ell^*$}
& Unified anchor $\mathbf{t}_{i}$
& Disjoint targets $\mathbf{t}^{\mathrm{bio}}_i \neq \mathbf{t}^{\mathrm{qa}}_i$ \\
 

\textbf{3. Representation at $\ell^*$}
& $\mathbf{h}^{\mathrm{bio}}_i \approx \mathbf{h}^{\mathrm{qa}}_i \approx \mathbf{t}_{i}$ (consistent)
& $\mathbf{h}^{\mathrm{bio}}_i \approx \mathbf{t}^{\mathrm{bio}}_i \neq \mathbf{h}^{\mathrm{qa}}_i \approx \mathbf{t}^{\mathrm{qa}}_i$ (inconsistent) \\
 
\textbf{4. Generalization}
& \makecell[l]{Unseen QA maps to seen BIO position\\ $\to$ successful recall}
& \makecell[l]{Unseen QA has no matching BIO position\\ $\to$ failed recall} \\
\bottomrule

\end{tabular}
}
\end{table*}

\section{Why Consistency Emerges: Optimization Dynamics}
\label{sec:dynamic_consistency}
Section~\ref{sec:role_of_representations} established that representation 
consistency drives recall success, but why does this consistency 
emerge in mixed training but not two-stage training? We approach this 
question from the perspective of optimization dynamics.

\zy{\textbf{Optimization Target}.}
Building on the observation that deep Pre-LN Transformer layers exhibit representation stability~\cite{sun2025the}, i.e.,  $\mathbf{h}_{\ell^*+1} \approx \mathbf{h}_{\ell^*} \approx \mathbf{h}_L$ (where $L$ is the total 
number of layers), we model the decoding process from layer $\ell^*$ as
\begin{equation}
P(a | \mathbf{h}_{\ell^*}) \approx \mathrm{Softmax}(\mathbf{h}_{\ell^*}\mathbf{W}_{\mathrm{unemb}}),
\label{eq:fixed_decoder}
\end{equation}
where $\mathbf{W}_{\mathrm{unemb}} \in \mathbb{R}^{d \times |\mathcal{V}|}$ projects the hidden state to the vocabulary space.
\zy{Then optimizing the BIO and QA tasks reduces to optimizing $\mathbf{h}_{\ell^*}$ to minimize the corresponding losses.}
For any fact $i$, we define $\mathbf{t}^{\mathrm{bio}}_i$ and $\mathbf{t}^{\mathrm{qa}}_i$ as the optimal $\mathbf{h}_{\ell^*}$ that minimize the BIO and QA losses, respectively.

\zy{
\textbf{Optimization Objective and Hypothesis.}
The two paradigms differ in their optimization objective. Two-stage training minimizes the BIO and QA losses sequentially: BIO first, then QA.
Mixed training minimizes a joint loss over BIO and QA simultaneously.}
We thus hypothesize that in two-stage training, without cross-task optimization constraints, each loss develops its own target, leaving them disparate in the latent space ($\mathbf{t}^{\mathrm{bio}}_i \ne \mathbf{t}^{\mathrm{qa}}_i$).
In mixed training, the joint loss pressures the model toward a unified solution under parameter budgets, leading to $\mathbf{t}^{\mathrm{bio}}_i = \mathbf{t}^{\mathrm{qa}}_i$.
We formalize our core hypothesis as follows:
\begin{tcolorbox}[
    colback=gray!5, 
    colframe=black!75, 
    arc=1mm, 
    boxrule=0.5pt, 
    left=2mm, right=2mm, top=2mm, bottom=2mm,
    title=\textbf{Hypothesis 2 (Optimal Target Consistency)}
]
\small
\vspace{1mm}
\begin{itemize}[leftmargin=4mm, labelsep=1mm, nosep]
\item \textbf{Mixed Training}: under the joint loss, $\mathbf{h}^{\mathrm{bio}}_{i}, \mathbf{h}^{\mathrm{qa}}_{i}$ converge  toward a unified anchor $\mathbf{t}_{i}$. This is the source of the representation consistency in H1.
\item \textbf{Two-stage Training}: under successive losses, $\mathbf{h}^{\mathrm{bio}}_{i}$, $\mathbf{h}^{\mathrm{qa}}_{i}$ converge toward separate targets ($\mathbf{t}^{\mathrm{bio}}_i \ne \mathbf{t}^{\mathrm{qa}}_i$).
\end{itemize}
\end{tcolorbox}

\textbf{Gradient Metric.} To test H2, we track gradient consistency between BIO and QA throughout training.
We define gradient consistency at each step as the cosine similarity between gradients projected into the recall-relevant subspace $\mathbf{U}$:
\begin{equation}
\mathrm{Cons}_{\mathrm{grad}}(\mathbf{g}_i^{\mathrm{qa}}, \mathbf{g}_i^{\mathrm{bio}}) = \frac{\langle \mathbf{g}_i^{\mathrm{qa}} \mathbf{U}, \mathbf{g}_i^{\mathrm{bio}} \mathbf{U} \rangle}{\|\mathbf{g}_i^{\mathrm{qa}} \mathbf{U}\| \|\mathbf{g}_i^{\mathrm{bio}} \mathbf{U}\|},
\end{equation}
where $\mathbf{g}_i^{\mathrm{qa}}$ and $\mathbf{g}_i^{\mathrm{bio}}$ denote the gradients of $\mathbf{h}_{i}^{\mathrm{qa}}$ and $\mathbf{h}_{i}^{\mathrm{bio}}$ at layer $\ell^*$ with respect to the QA and BIO losses, respectively. $\mathbf{U}$ is the same recall-relevant subspace used for representation consistency in $\S$~\ref{sec:role_of_representations}.

Since gradients determine how representations are updated at each step, jointly analyzing gradient consistency and representation consistency reveals the target of convergence during training.
For example, consistent gradients together with consistent representations throughout training indicate convergence toward a unified target; while inconsistent gradients with inconsistent representations indicate separate targets.

\textbf{Experimental setup.} 
We compute $\mathrm{Cons}_{\mathrm{grad}}$ on 5k individuals from the out-of-distribution set, and approximate the training trajectory from saved checkpoints.
Because $a_4$ exhibits a clear gap in representation consistency and recall between mixed and two-stage training, we use it as a meaningful example.
Appendix~\ref{appendix:details_of_emerged_consistency} shows  trends for $a_1$ and $a_6$.

\begin{figure}
    \centering
    \includegraphics[width=1\linewidth]{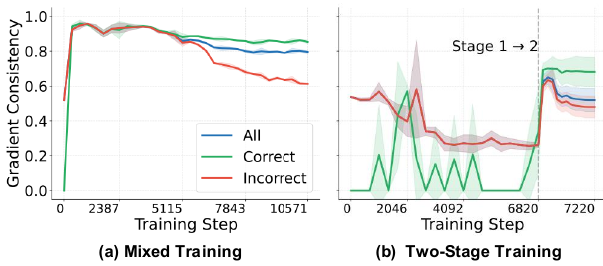}
    \caption{Each panel shows gradient consistency between BIO and QA tasks across training.
    Mixed training maintains aligned gradients throughout training, while two-stage training does not.
    }
    \label{fig:llama_gradient_evolution}
\end{figure}

\textbf{Gradient Results.}
Figure~\ref{fig:llama_gradient_evolution} shows that the two strategies exhibit opposite gradient dynamics throughout training.
Since representation consistency tracks gradient consistency closely (Appendix~\ref{appendix:details_of_emerged_consistency}), consistent gradients throughout training indicate continuous convergence toward a unified target, while inconsistent gradients indicate convergence toward separate targets.
In mixed training, the \textit{All} curve closely tracks the \textit{Correct} curve, indicating that most samples are correctly recalled.
These correctly recalled samples maintain high gradient consistency ($\sim$0.8--1.0) throughout training, showing that the joint loss continually aligns BIO and QA gradients and drives most samples toward a unified anchor $\mathbf{t}_i$.
Two-stage training shows the opposite pattern. 
Specifically, during training, the \textit{All} curve stays close to the \textit{Wrong} group, indicating that most samples fail recall.
In Stage-1 (BIO loss), gradients between BIO and QA samples diverge sharply.
At the start of Stage-2 (QA loss), gradient consistency jumps but remains substantially lower than in mixed training (below 0.7), indicating that BIO and QA representations continue to be driven toward separate targets ($\mathbf{t}^{\mathrm{bio}}_i \ne \mathbf{t}^{\mathrm{qa}}_i$). 
Similar behaviors in Pythia and Qwen models are provided in Appendix~\ref{appendix:details_of_emerged_consistency}.

\zy{\textbf{Where Consistency Resides.}}
\zy{The BIO-QA gradient consistency also leaves a measurable trace in parameter space.}
\zy{By the chain rule $\nabla_{\theta_{\ell}}= \left(\frac{\partial \mathbf{h}_{\ell+1}}{\partial \theta_{\ell}}\right)^\top \nabla_{\mathbf{h}_{\ell+1}}$, gradient consistency gives rise to a set of shared parameters that receive substantial updates from both BIO and QA tasks.}
\zy{Although these parameters constitute only $\sim$9.3\% of the recall-relevant parameters, they contribute over 59\% of model total performance (Figure~\ref{fig:shared_params_chart} in Appendix~\ref{appendix:details_of_shared_parameters}), acting as its mechanistic core.}
\zy{Mixed training establishes a substantially larger pool of such parameters than two-stage training (58.2\% Acc.\ / 0.47M vs.\ 27.6\% Acc.\ / 0.38M on Llama), revealing a structural foundation that two-stage training fails to build.}

\begin{table*}[t]
\centering
\caption{Four input variants of the BIO sequence ($s, r_1, a_1, \dots, r_4, a_4$), e.g., \emph{Meghan Charles's birthday is remembered on November 10, 2092. He took birth in Amyville, GA. He took advantage of the diverse programs offered at Cedar Crest College. He was involved in work as a chartered public finance accountant.} We show example when predicting attribute $a_4$: ``chartered public finance accountant.''}
\resizebox{1\linewidth}{!}{
\begin{tabular}{p{0.2\textwidth} p{0.25\textwidth} p{0.1\textwidth} p{0.45\textwidth}}
\toprule
\rowcolor{headercolor}
\textbf{Variant} & \textbf{Description} & \textbf{Input} & \textbf{Input Example} \\
\midrule
\textbf{Subject} & Only the subject name
& $s$
& Meghan Charles \\
\hline
\textbf{Full w/o Relation} & Preceding context truncated before the target relation
& $s, r_1, a_1,$ $ \dots, r_{3}$
& Meghan Charles's birthday is remembered on November 10, 2092. He took birth in Amyville, GA. He took advantage of the diverse programs offered at \\
\hline
\textbf{Subject + Relation} & The subject name and target relation
& $s, r_4$
& Meghan Charles was involved in work as \\
\hline
\textbf{Full} & Full preceding tokens
& $s, r_1, a_1,$ $ \dots, r_4$
& Meghan Charles's birthday is remembered on November 10, 2092. He took birth in Amyville, GA. He took advantage of the diverse programs offered at Cedar Crest College. He was involved in work as a \\
\bottomrule
\end{tabular}
}
\label{tab:input-levels}
\end{table*}

\subsection{Why Does Mixed Training Work?}
Table~\ref{tab:comparison} summarizes the fundamental differences between mixed and two-stage training.
\zy{Mixed training succeeds because its joint optimization signal binds 
storage and query into a format-invariant retrieval mechanism.}

\zy{The mechanism unfolds as a causal chain established across the preceding 
sections.}
\zy{By optimizing BIO and QA samples jointly, mixed training produces \emph{gradient consistency} across the two formats.}
\zy{It aligns each individual's BIO hidden 
state with its QA counterpart, producing \emph{representation consistency}
($\mathbf{h}^{\mathrm{bio}}_i \approx \mathbf{h}^{\mathrm{qa}}_i \approx \mathbf{t}_i$, \S\ref{sec:dynamic_consistency}).}
\zy{The \emph{representation consistency} establishes a retrieval mechanism that maps unseen QA queries to facts stored during BIO training, regardless of surface format, thereby enabling successful recall (\S\ref{sec:role_of_representations}).}

\zy{Two-stage training lacks the joint objective. With BIO and QA gradients arriving in disjoint phases, the model receives no direct signal to produce gradient consistency, leading to representations inconsistent across formats. As a result, although the same factual content is stored, 
QA queries are unable to retrieve it.
}

\zy{This contrast reframes the distinction between the two paradigms: mixed training does not store more facts (\S\ref{section:preliminary_experiments});
rather, its stored facts exhibit format-invariant representations across BIO and QA, making them retrievable from QA queries. This format-invariance is what distinguishes genuine knowledge from surface-level memorization.}

\begin{figure*}[t]
    \centering
    \includegraphics[width=1\linewidth]{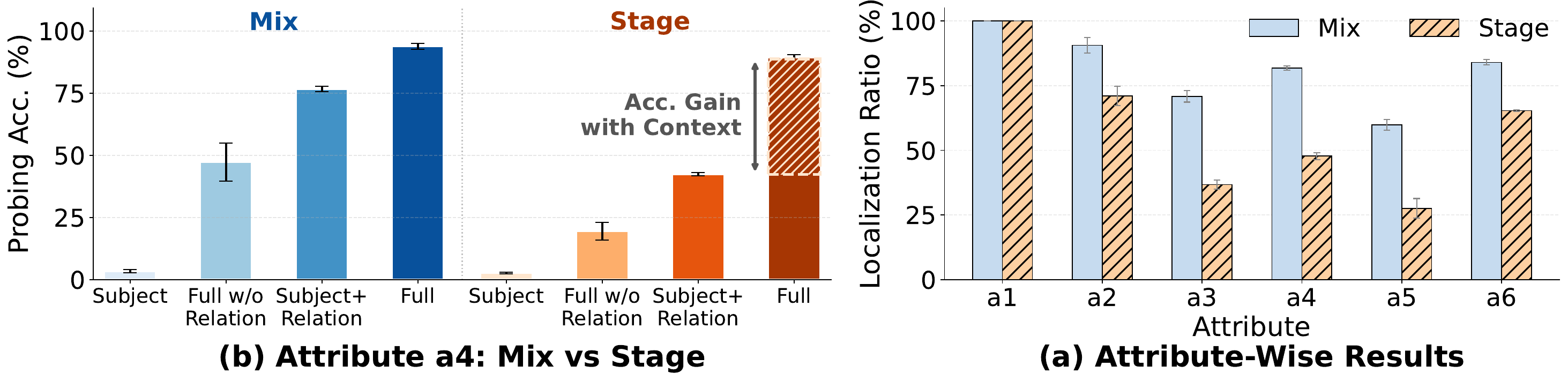}
    \caption{
    \textbf{(a)} Accuracy of the four input variants. 
    \textbf{(b)} Localization ratio across attributes.
    Mix-tuned Llama encodes attributes in the BIO format through \emph{Subject + Relation}, while stage-tuned Llama relies on the \emph{Full} preceding context.
    Pythia and Qwen models exhibit similar behavior, as detailed in Appendix~\ref{appendix:pythia_bio_attribute_encoding}.
    }
    \label{fig:bio_saving_pos_bar}
\end{figure*}

\section{How Consistency Appears at the Input}
\label{sec:local_grounding}
We have shown that mixed training succeeds through representation consistency (\S\ref{sec:role_of_representations}), driven by the joint training objective (\S\ref{sec:dynamic_consistency}).
This mechanism operates in the latent space. We now ask whether it leaves an interpretable signature at the input level: which input tokens does the model rely on to encode attributes in the storage format.
Prior work identified the subject token as a primary anchor \citep{pmlr-v235-allen-zhu24a}. 
\zy{We further hypothesize that the relation token is also essential.
The reasoning follows from our earlier findings: since queries contain only the subject and target relation, consistent storage and query representations require the model to encode attributes from the subject and target relation alone.}
As listed in Table~\ref{tab:input-levels}, we design four input variants to test this.
The relation token is essential if variants with it (\textit{Subject + Relation} and \textit{Full}) substantially outperform those without (\textit{Subject} and \textit{Full w/o Relation}).
Moreover, a model with consistent representations should encode attributes from the subject and target relation alone, so its \textit{Subject + Relation} performance should approach \textit{Full}.

\textbf{Experimental Setup.} For each input variant, we freeze the fine-tuned model's parameters and train a linear classifier on the final hidden layers to predict the target attribute. 
We use 9k BIO samples for training and 1k for evaluation (Appendix~\ref{appendix:pythia_bio_attribute_encoding}).
We define the localization ratio as ($\mathrm{Acc}(Subject+Relation)/\mathrm{Acc}(Full)$) to quantify how closely \textit{Subject + Relation} approaches \textit{Full}.

\textbf{Relation is an Essential Anchor.}
Figure~\ref{fig:bio_saving_pos_bar}a shows that variants with the relation (\textit{Subject + Relation}, \textit{Full}) substantially outperform those without (\textit{Subject}, \textit{Full w/o Relation}).
This confirms that the target relation is an essential anchor for encoding the attribute.

\textbf{Encoding Results.} 
Figure~\ref{fig:bio_saving_pos_bar}b shows that the mix-tuned model achieves a high localization ratio ($\ge$56\%) across attributes.
The stage-tuned model, which suffers from representation inconsistency, exhibits a much lower ratio (e.g., 47.8\% vs.\ 81.6\% on $a_4$).
\zy{This comparison reflects representation consistency at its source. Mixed training encodes attributes in storage format from the subject and relation, which are the same components a query provides, so the storage and query representations are built from a shared input basis.
Two-stage training predominantly encodes from the full preceding context, which a query lacks, leaving the two representations without a shared input basis.}

\section{Related work}
\textbf{Mechanistic interpretability} aims to explain model behavior by uncovering its internal mechanisms~\cite{gurnee2024universal,marks2024the,voita-etal-2024-neurons,zhao2024how}.
Prior work has investigated how MLP neurons~\cite{gurnee2024universal,gurnee2023finding,wang-etal-2022-finding-skill} and attention heads~\cite{prakash2024finetuning,zhao2024how} contribute to particular linguistic or reasoning tasks.
Other studies have explored whether models develop internal capabilities aligned with human-interpretable concepts, such as geometry~\cite{gurnee2024language}, monotonicity~\cite{heinzerling-inui-2024-monotonic}, or linearity~\cite{engels2025not,hernandez2024linearity}.
Various tools, including the logit lens~\cite{nostalgebraist2020logit} and path patching~\cite{zhang2024towards}, have also been developed.
Recent work has analyzed how different optimization strategies affect a model’s internal capabilities~\cite{jain2024mechanistically,petrov2024when}, the underlying driving forces behind these evolutions remain less understood. We extend this by introducing a gradient-based perspective to explain the formation of factual knowledge.

\textbf{Instruction-tuning}~\cite{NEURIPS2022_b1efde53} fine-tunes LMs on (INSTRUCTION, OUTPUT) pairs.  
We adopt this setup for QA-tuning.
Subsequent work has explored alternative data sources~\cite{longpre2023flan,wang-etal-2022-super,zhou2023lima}, tuning strategies~\cite{bai2022training,brooks2023instructpix2pix,cheng-etal-2024-instruction}, tuned-LMs~\cite{DatabricksBlog2023DollyV2,dai2023instructblip}, and evaluation~\cite{adlakha-etal-2024-evaluating,ChungHLZTFL00BW24}.
%
%
We focus on interpreting the mechanisms behind different instruction-tuning strategies. By comparing sequential and mixed training, we uncover why joint optimization leads to superior knowledge internalization.

\textbf{Understanding factual knowledge in language models} falls into two main directions: understanding how models \textit{store} facts~\cite{dai-etal-2022-knowledge,heinzerling-inui-2021-language,heinzerling-inui-2024-monotonic,katz-etal-2024-backward,pmlr-v202-maini23a,yao2024knowledge,pmlr-v235-allen-zhu24a}, and how they \textit{recall} facts~\cite{allen-zhu2025physics,geva-etal-2023-dissecting,ghosal2024understanding,ortu-etal-2024-competition,choe-etal-2025-autoregressive}.
\zy{Most relevant to our work, \citet{pmlr-v235-allen-zhu24a} contrast the recall behaviors of two-stage and mixed training, highlighting the role of subject tokens in encoding attributes.}
\zy{We extend this framework in two critical dimensions. First, we bridge the gap between training paradigms and recall behavior by providing a mechanistic explanation. Second, we demonstrate that beyond subject tokens, relation tokens are also an essential component for encoding.}
\zy{While prior work suggests storage and retrieval may share similar mechanisms~\citep{yao2024knowledge,geva-etal-2023-dissecting,ghosal2024understanding}, we provide the missing mechanistic link by demonstrating that representation consistency is the prerequisite for successful recall. Facts are only reliably retrieved when their storage and query prompts converge toward a unified internal anchor during optimization.}

\section{Conclusion}
\zy{Through representation and gradient analyses,
we show that mixed training enables fact recall by establishing format-invariant representations between storage and query.}
\zy{This mechanism provide a foundation for knowledge injection: effective fine-tuning should encourage the representations of two formats to align, as mixed training does, rather than present them sequentially.}
\zy{While we focus on interpreting fact recall, this analysis framework provides three diagnostic metrics (geometric, behavioral, and gradient) applicable to broader questions of knowledge integrity.}

\section*{Limitations}
\label{appendix:limitation}
While our work offers a strong empirical foundation for understanding the effect of mixed training on fact recall, several limitations remain for future exploration:
\textbf{(1) Model Scope.} Our analysis covers three text-only LMs with up to 4 billion parameters.
Whether the observed patterns generalize to larger or multimodal models, such as those incorporating image or video inputs, remains open.
\textbf{(2) Data Realism.}
Our datasets are synthetic, clean, balanced, and by construction do not conflict with the model's pretrained knowledge.
Real-world knowledge injection often involves noise, imbalance, and information that conflicts with or overrides existing knowledge. Recall under such conditions is beyond our current scope.
\textbf{(3) Scope of Fact Recall.} We focus on facts injected during fine-tuning, not knowledge already acquired during pretraining. The recall of pretrained knowledge is not addressed here.
\textbf{(4) Theoretical Assumptions.} Our analysis builds on two assumptions from prior work. 
First, retrieval forms an internal representation from the subject and relation before extracting the attribute~\citep{geva-etal-2023-dissecting}, this motivates our focus on a specific layer $\ell^*$.
Second, attribute extraction can be modeled as a linear readout~\citep{hernandez2024linearity}, this underlies our use of a fixed readout matrix $\mathbf{W}^{\mathrm{qa}}$ to attribute recall failure to representation inconsistency.
Cases where these assumptions do not hold are beyond the scope of this work.

\section*{Acknowledgment}
We thank Yuta Hitomi and Ryoma Ishigaki for insightful discussions.

\bibliography{latex/custom}

\appendix

\appendix

\section{Appendix / Supplemental Material}
\subsection{Details on Data Preparation}
\label{appendix:data_preparation}
\subsubsection{BIO Dataset}
This research followed the general setup of~\cite{pmlr-v235-allen-zhu24a} with some modifications to generate synthetic datasets. Specifically, we generated profiles for $N=10,000$ individuals. For each person, we independently and randomly sampled their first and last name, gender, birth date, birth city, attended university, job, employer, and blood type from uniform distributions. Details are as follows:
\begin{itemize}
    \item First and last names were drawn from pools of 400 and 1000 English names, respectively. Each of the 10,000 individuals had a unique full names to ensure disambiguation during fact recall.
    \item Birth dates were generated with years from 1900 to 2099, months from 1 to 12, and days from 1 to 28.
    \item Birth cities were selected from a pool of 200 fake cities, each randomly assigned to one of 56 U.S. state abbreviations. Birth cities were formatted as, e.g., Amyville, GA.
    \item Attended Universities were selected from a list of 300 U.S. higher education institutions.
    \item Jobs were draw from a pool of 100 occupations (e.g., chartered public finance accountant).
    \item Employers were selected from a pool of 263 fake Companies (e.g., Bell, Ramos and Romero).
    \item Blood types were sampled from the standard 8 blood types, including A+, A-, B+, B-, AB+, AB-, O+, O-.
    \item Genders was selected as either F or M.
\end{itemize}
We used the \texttt{Faker} Python module to generate the pools of names, cities, jobs, and employers. For each individual, we generated one biographical entry consisting of six sentences, each corresponding to one attribute, and the entry maintained a consistent order: birth date, birth city, attended university, job, employer, and blood type. Each sentence was instantiated from a randomly selected template.
Specifically, we used 21 templates for birth dates, 33 for birth cities, 22 for universities, 24 for jobs, 29 for employers, and 21 for blood types. Pronouns (she/he) were assigned based on the individual's gender (F/M).
Example biographies:
\begin{itemize}
    \item \textit{Brandon Horne's life journey began on February 1, 2089. She spent her youth in Cindyborough, AL. She developed a passion for learning at University of North Texas Health Science Center at Fort Worth. She had embraced a career as a landscape architect. She accepted a leadership position at Bell, Ramos and Romero. She is classified as having blood AB+.}
    \item \textit{Meghan Charles's birthday is remembered on November 10, 2092. He took birth in Amyville, GA. He took advantage of the diverse programs offered at Cedar Crest College. He was involved in work as a chartered public finance accountant. He aligned his professional ambitions with Henson, Ellis and Sexton. He has a blood type of O-.}
\end{itemize}
To evaluate BIO accuracy, we used part of the biography as a prompt and asked the model to generate the corresponding attribute. Accuracy was measured as the proportion of exact matches. Example:
\begin{itemize}
    \item Prompt: Brandon Horne's life journey began on February 1, 2089. She spent her youth in
    \item Attribute: Cindyborough, AL.
\end{itemize}

\subsubsection{QA Dataset}
For each individual, we generated six fixed questions corresponding to the six attributes. The question served as the prompt, and the model generated the attribute as the answer. Accuracy was again computed by exact match. Examples:
\begin{itemize}
    \item \textit{What is the birth date of Meghan Charles? November 10, 2092.}
    \item \textit{What is the birth city of Meghan Charles? Amyville, GA.}
    \item \textit{Which university did Meghan Charles study? Cedar Crest College.}
    \item \textit{What is the job of Meghan Charles ? chartered public finance accountant.}
    \item \textit{Which company did Meghan Charles work for? Henson, Ellis and Sexton.}
    \item \textit{What is the blood type of Meghan Charles ? O-.}
\end{itemize}

\subsection{Details on BIO/QA Fine-Tuning}
\label{appendix:bioqa_finetuning}

\begin{table}[h]
\centering
\caption{Training updates for different model and training strategies.}
\resizebox{1\linewidth}{!}{
\begin{tabular}{llrr}
\toprule
\multirow{2}{*}{\textbf{Model}} & \multirow{2}{*}{\textbf{Training Type}} & \multicolumn{1}{l}{\multirow{2}{*}{\textbf{Updates}}} & \multicolumn{1}{l}{\multirow{2}{*}{\textbf{Training Hours (1 run)}}}\\
                                &                                    & &\multicolumn{1}{l}{}                                  \\ \midrule
\multirow{3}{*}{Llama}          & BIO                                & 6,820                           &       10.37               \\
                                & QA                                 & 400        &  4.36                                         \\
                                & Mix                                & 10,571       &     22.16                                    \\ \midrule
\multirow{3}{*}{Pythia}         & BIO                                & 13,981                                   &    18.27         \\
                                & QA                                 & 800          &    6.15                                     \\
                                & Mix                                & 19,437       &    32.43                             \\ \midrule
\multirow{3}{*}{Qwen}         & BIO                                &              4,928                       &    1.87         \\
                                & QA                                 &     400      &     0.88                                    \\
                                & Mix                                &    10,304    &     3.02     \\ \bottomrule

\end{tabular}
}
\label{table:num_bioqa_updates}
\end{table}

\begin{figure*}[h]
    \centering
    \includegraphics[width=1\linewidth]{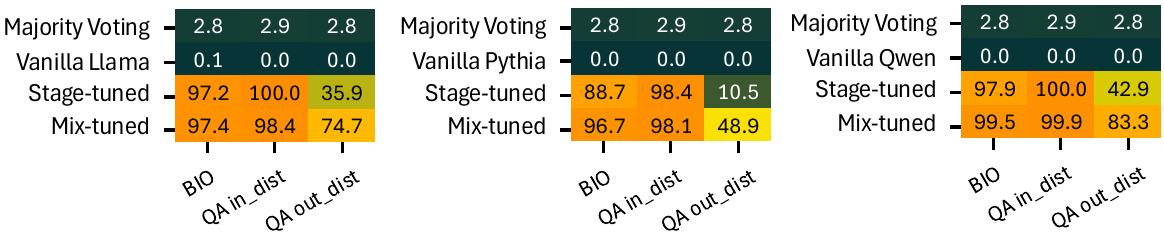}
    \caption{Performance of fine-tuned Llama and Pythia on BIO, QA in-distribution, and QA out-of-distribution sets.}
    \label{fig:finetuned_llm_performance}
\end{figure*}

\textbf{Hyperparameters.} We fine-tuned all models using the AdamW optimizer with an initial learning rate of 0.0001 and cosine learning rate decay on NVIDIA A100 80GB. All models were trained with next-token prediction. For BIO fine-tuning and mixed training, we used a linear warm-up of 1,600 steps and a batch size of 32. For QA fine-tuning, we used no warm-up and a batch size of 256.
Table~\ref{table:num_bioqa_updates} summarizes the total number of training updates for each model setting. The estimated the total GPU compute time in fine-tuning models is 281.22 hours without taking preliminary or failed experiments (that didn’t make it into the paper) into consideration.

\begin{figure}[h]
    \centering
    \includegraphics[width=1\linewidth]{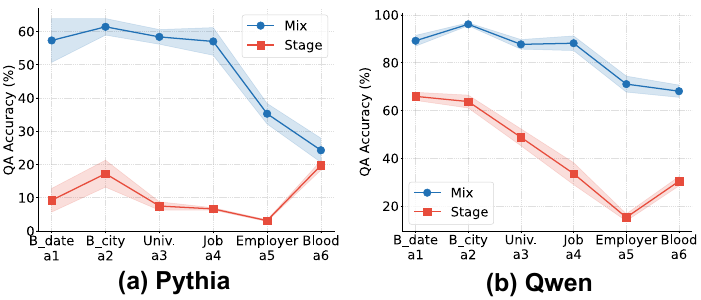}
    \caption{Per-attribute accuracy of Pythia and Qwen models.}
    \label{fig:qwen_pythia_per_attr}
\end{figure}

Figure~\ref{fig:finetuned_llm_performance} shows the performance of fine-tuned models. Both the QA-tuned and Mix-tuned models perform well on data seen during training. Specifically, they retain biographical facts (BIO Accuracy $\ge$ 88.7\%) and adapt well to the question-answer format (QA in\_dist accuracy $\ge$ 98.1\%).
Figure~\ref{fig:qwen_pythia_per_attr} shows the per-attribute QA accuracy of Pythia and Qwen models.

\subsection{Details on Representation Consistency}
\label{appendix:details_relation_representation}
We formalize the extraction process as a linear probe $\mathbf{W} \in \mathbb{R}^{d \times |\mathcal{V}|}$, where $P(a | \mathbf{h}) \approx \mathrm{Softmax}(\mathbf{h}\mathbf{W})$ estimates the probability of the first token of the target attribute. 
To improve the robustness of the probe, we decompose $\mathbf{W}$ into $\mathbf{W} = \mathbf{W}_r \mathbf{W}_\mathrm{unemb}$, where $\mathbf{W}_r \in \mathbb{R}^{d \times d}$ is a learned projection matrix that maps hidden states $\mathbf{h}$ to the final-layer representation space, and $\mathbf{W}_{\mathrm{unemb}} \in \mathbb{R}^{d \times |\mathcal{V}|}$ is the trained model's fixed unembedding layer.
We optimize $\mathbf{W}_r$ using a multi-task objective:
\begin{align}
\mathcal{L} = &\mathcal{L}_{\mathrm{MSE}}(\mathbf{h}\mathbf{W}_r, \mathbf{h}_{\mathrm{final}}) \nonumber \\ + &\mathcal{L}_{\mathrm{CE}}(\mathrm{Softmax}(\mathbf{h}\mathbf{W}), a),
\end{align}
where $\mathcal{L}_{\mathrm{MSE}}$ and $\mathcal{L}_{\mathrm{CE}}$ denote mean squared error and cross-entropy loss, respectively.

\textbf{Hyperparameters.}
We employ the AdamW optimizer with an initial learning rate of 0.001 and a linear decay schedule (no warm-up). The batch size is set to 100. We perform 2,000 training updates for Pythia models and 1,500 updates for Llama models to ensure convergence.
Notably, since Pythia utilizes a parallel attention and MLP architecture, we define its $\mathbf{h}$ as the total output of layer $\ell^*$.
We define $\ell^*=30$ for Pythia models and $\ell^*=33$ for Qwen models, since $\mathrm{Acc}(\mathbf{h}^{\mathrm{qa}}\mathbf{W}^{\mathrm{qa}})$ saturates in subsequent layers in  Figures~\ref{fig:pythia_incosistency_induced_recall}a,d.

\subsubsection{Optimization for Matrix Decomposition}
\label{appendix:matrix_decomposition}
We solve for $\mathbf{U}$ by optimizing a multi-objective function, formulated as:
\begin{align}
& \min_{\mathbf{U},\ \mathbf{H}^{\text{qa}}_{\text{attr}},\ \mathbf{H}^{\text{bio}}_{\text{attr}}} \nonumber \\
\ \ 
&\underbrace{\|\mathbf{H}^{\text{qa}}-\mathbf{H}^{\text{qa}}_{\text{attr}}\mathbf{U}^\top\|_F^2
+
\|\mathbf{H}^{\text{bio}}-\mathbf{H}^{\text{bio}}_{\text{attr}}\mathbf{U}^\top\|_F^2}_{\text{Reconstruction Loss}} \nonumber  \\ 
& \quad \;-\;
 \alpha\,
\underbrace{\frac{\mathrm{tr}(\mathbf{U}^\top C_\mathcal{C} \mathbf{U})}{\mathrm{tr}\!\big(\mathbf{U}^\top(C_\mathcal{W}+\varepsilon I)\mathbf{U}\big)}}_{\text{Discriminative Penalty}} \nonumber \\ 
& \text{s.t.} \quad \mathbf{U}^\top \mathbf{U} = \mathbf{I},\quad \mathbf{U}\mathbf{U}^\top \mathbf{W}^{\mathrm{qa}}=\mathbf{W}^{\mathrm{qa}}.
\end{align}
This objective consists of two primary components.
First, the \textit{Reconstruction Loss} identifies a shared subspace $\mathbf{U}$ that spans both BIO and QA representations.
Second, the \textit{Discriminative Penalty} ensures that recall failure is sensitive to the representation inconsistency.
Specifically, this term utilizes the variance of representation differences, $\Delta \mathbf{h}_i = \mathbf{h}_i^{\text{qa}} - \mathbf{h}_i^{\text{bio}}$, across different recall outcomes. 
Let $\mathcal{C}$ and $\mathcal{W}$ be the sets of samples where the probe $\mathbf{W}^{\text{qa}}$ yields correct and incorrect predictions, respectively.
We define:
\begin{equation}
C_\mathcal{K} = \sum_{i \in \mathcal{K}} (\Delta \mathbf{h}_i - \boldsymbol{\mu}_\mathcal{K}) (\Delta \mathbf{h}_i - \boldsymbol{\mu}_\mathcal{K})^\top, 
\end{equation}
where $\boldsymbol{\mu}_\mathcal{K}$ is the mean difference vector within group $\mathcal{K} \in \{\mathcal{C}, \mathcal{W}\}$.
The \textit{Discriminative Penalty} minimizes the projected variance of $\Delta \mathbf{h}_i$ for correct samples ($C_\mathcal{C}$) while maximizing it for incorrect ones ($C_\mathcal{W}$).
This ensures that the identified subspace captures the functional dimensions where inconsistency leads to recall failure.

\subsubsection{Validation of Probing Robustness}
\label{appendix:potential_bias}
To validate the proposed measure of representational consistency, we must ensure that our probing framework satisfies three critical constraints: format-invariance, syntactic invariance, and semantic specificity.

\begin{figure}[h]
    \centering
    \includegraphics[width=1\linewidth]{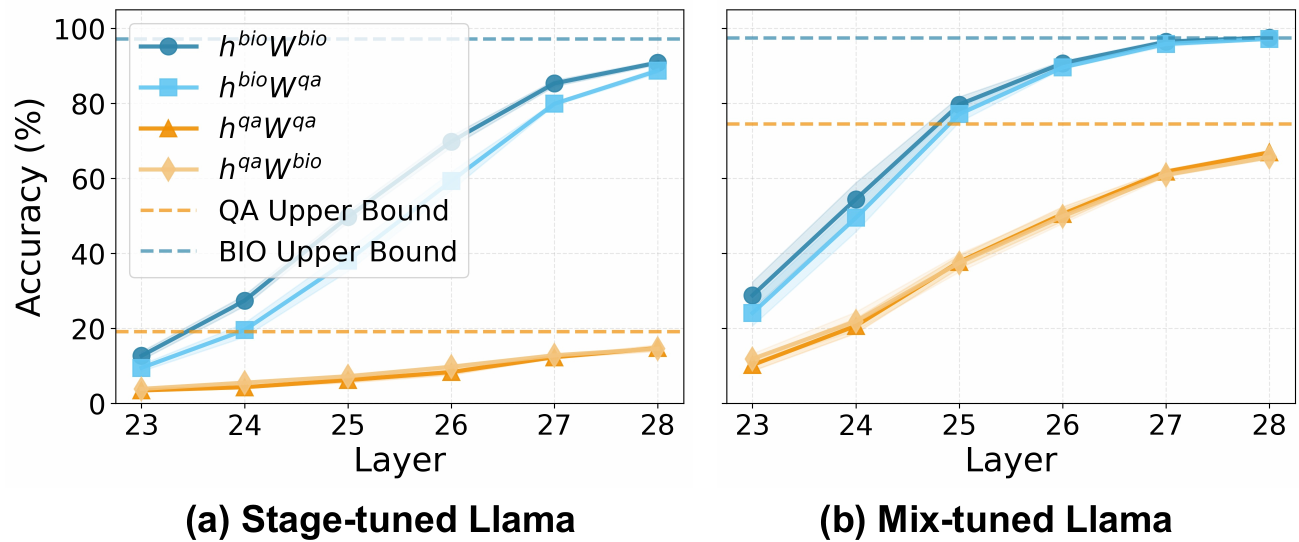}
    \caption{Identify format-invariance when predicting the first token of $a_4$.}
    \label{fig:llama_unified_probe}
\end{figure}

\paragraph{Format-Invariance.} 
We first verify whether the model employs a unified extraction logic across different input formats. We train two separate probes for attribute $a_4$, $\mathbf{W}^{\mathrm{bio}}$ (trained on BIO data) and $\mathbf{W}^{\mathrm{qa}}$ (trained on QA data).
Figure~\ref{fig:llama_unified_probe} shows that
$\text{Acc}(\mathbf{h}^{\mathrm{bio}}\mathbf{W}^{\mathrm{qa}}) \approx \text{Acc}(\mathbf{h}^{\mathrm{bio}}\mathbf{W}^{\mathrm{bio}})$ and 
$\text{Acc}(\mathbf{h}^{\mathrm{qa}}\mathbf{W}^{\mathrm{qa}}) \approx \text{Acc}(\mathbf{h}^{\mathrm{qa}}\mathbf{W}^{\mathrm{bio}})$.
This near-equivalence demonstrates that the extraction mechanism is robust to the format on which the probe was trained, confirming a consistent readout logic.

\begin{figure}[h]
    \centering
    \includegraphics[width=1\linewidth]{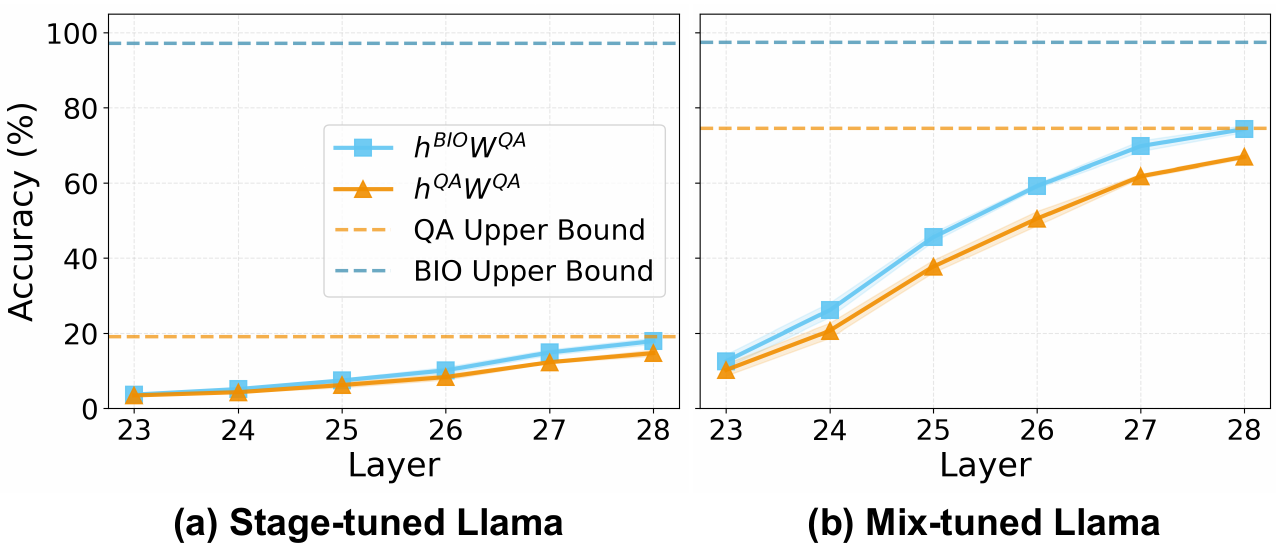}
    \caption{Identify syntactic invariance when predicting the first token of $a_4$.}
    \label{fig:llama_syntactic_invariance}
\end{figure}

\paragraph{Syntactic Invariance.}
We then examine if the surface-level difference between a statement (e.g., ``Subject's job is...'') and a question (e.g., ``What is the job of...?'') alters the underlying latent representation $\mathbf{h}$.
We compare the states derived from a short BIO trigger ($\mathbf{h}^{\mathrm{bio}}$) and a QA prompt ($\mathbf{h}^{\mathrm{qa}}$). Since both inputs convey the same subject-relation $(s, r)$ pair through different syntax, a gap in decoding accuracy would signal syntactic bias.
Figure~\ref{fig:llama_syntactic_invariance} confirms $\text{Acc}(\mathbf{h}^{\mathrm{bio}}\mathbf{W}^{\mathrm{qa}}) \approx \text{Acc}(\mathbf{h}^{\mathrm{qa}}\mathbf{W}^{\mathrm{qa}})$, ensuring that $\mathbf{h}$ is invariant to phrasing differences.

\begin{figure}[h]
    \centering
    \includegraphics[width=1\linewidth]{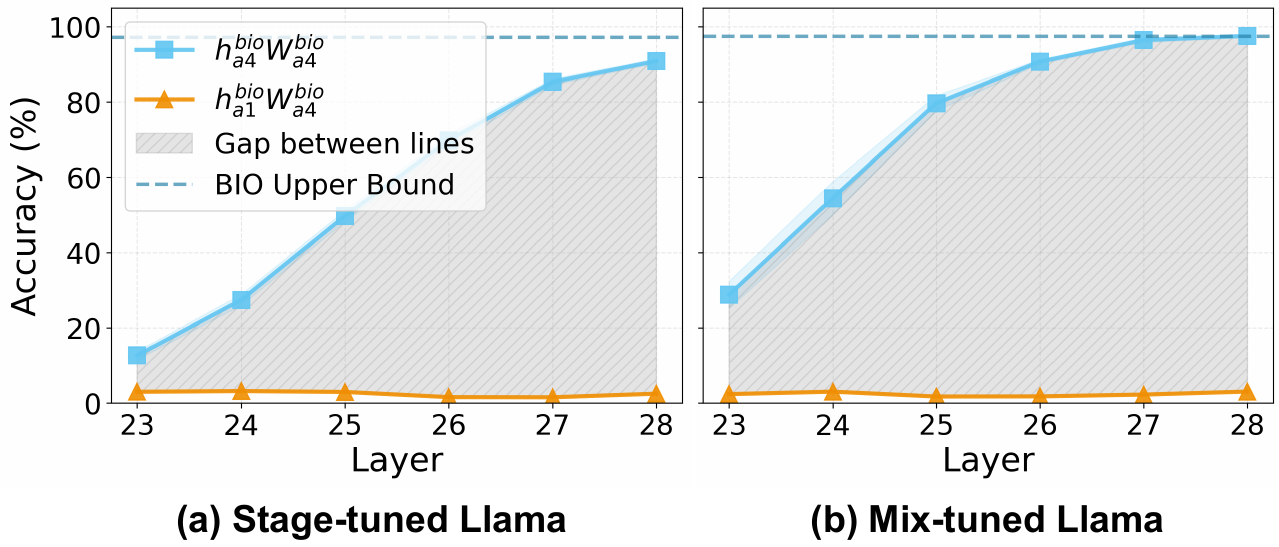}
    \caption{Identify semantic specificity when predicting the first token of $a_4$.}
    \label{fig:llama_semantic_specificity}
\end{figure}

\begin{figure*}[h]
    \centering
    \includegraphics[width=1\linewidth]{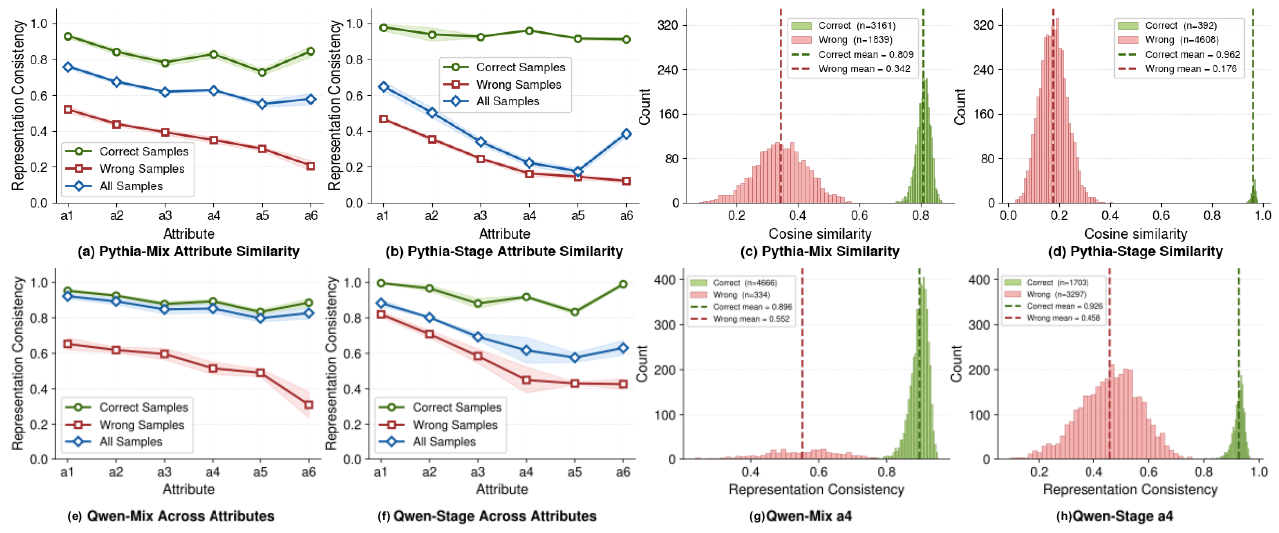}
    \caption{Quantitative analysis of representation consistency in Pythia and Qwen.}
    \label{fig:pythia_consistency}
\end{figure*}

\begin{figure}[h]
    \centering
    \includegraphics[width=1\linewidth]{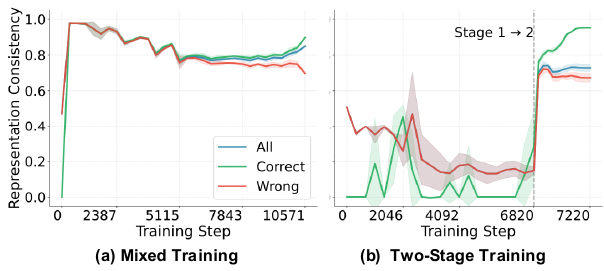}
    \caption{Evolution of representation consistency during training Llama.}
    \label{fig:llama_representation_evolution}
\end{figure}

\paragraph{Semantic Specificity.} 
Finally, we perform a cross-attribute swap test to ensure the probe captures attribute-specific information rather than generic patterns.
A probe $\mathbf{W}^{\mathrm{bio}}_{a_4}$ trained for a specific attribute (e.g., ``Job'') fails to decode its target information from hidden states $\mathbf{h}^{\mathrm{bio}}_{a_1}$ associated with a different attribute (e.g., ``Birthday'').
This significant performance gap, shown in Figure~\ref{fig:llama_semantic_specificity} as $\mathrm{Acc}(\mathbf{h}^{\mathrm{bio}}_{a_4}\mathbf{W}^{\mathrm{bio}}_{a_4}) \gg \mathrm{Acc}(\mathbf{h}^{\mathrm{bio}}_{a_1}\mathbf{W}^{\mathrm{bio}}_{a_4})$, confirms that $\mathbf{h}$ captures attribute-specific semantics rather than generic factual patterns.

\begin{figure*}[t]
    \centering
    \includegraphics[width=1\linewidth]{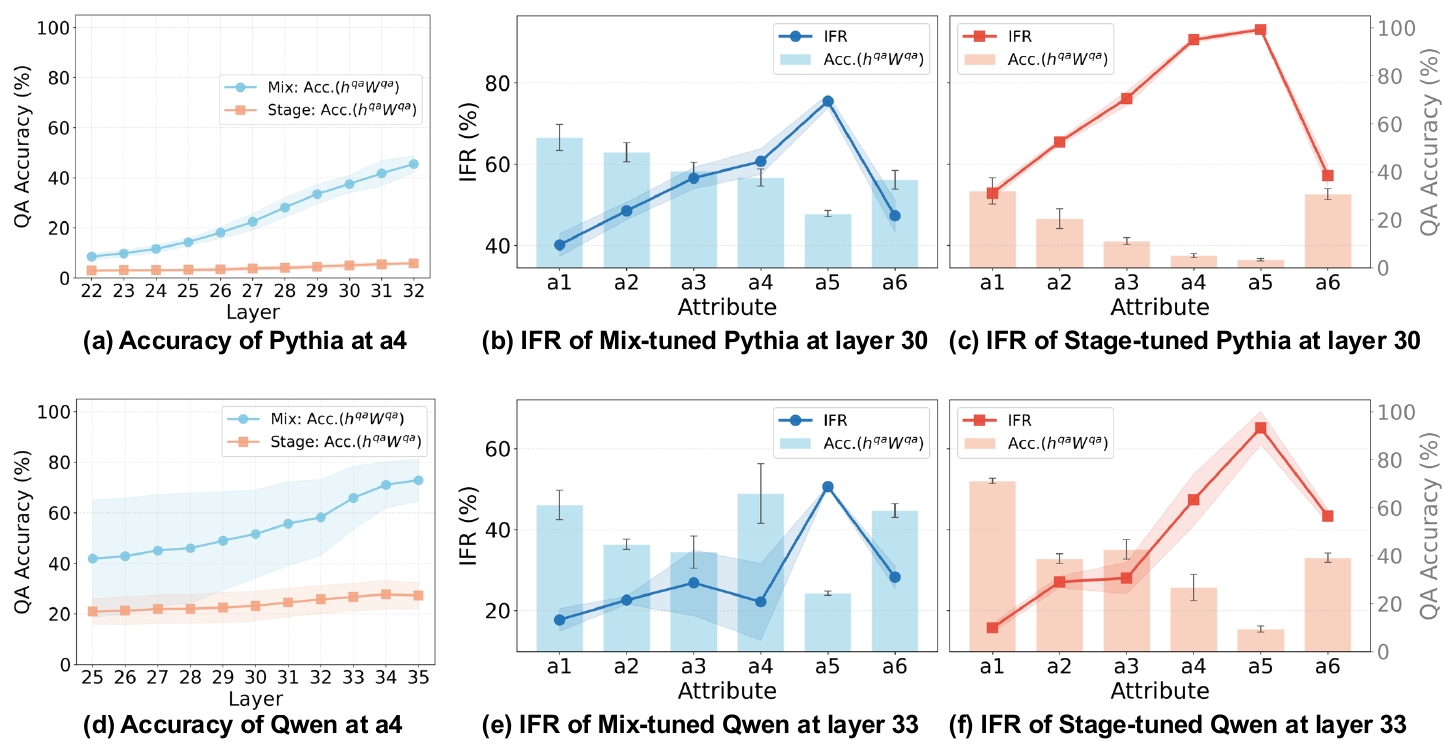}
    \caption{Impact of representation inconsistency on recall failure in Pythia and Qwen.} 
    \label{fig:pythia_incosistency_induced_recall}
\end{figure*}

\subsubsection{Results on Pythia and Qwen models.}
\label{appedix:representation_pythia_qwen}
Figures~\ref{fig:pythia_consistency} and \ref{fig:pythia_incosistency_induced_recall} present the results for Pythia and Qwen models.
Figure~\ref{fig:pythia_consistency} shows that, mixed training maintains the majority of samples within the high-consistency zone, whereas the most of samples in two-stage training fall into the low-consistency failure zone. 
As shown in Figures~\ref{fig:pythia_incosistency_induced_recall}b,c,e,f, stage-tuned models tend to have higher inconsistency-induced failure ratio than mix-tuned models.

\subsection{Details on Gradient Consistency}
\label{appendix:details_of_emerged_consistency}

\begin{figure*}[h]
    \centering
    \includegraphics[width=0.8\linewidth]{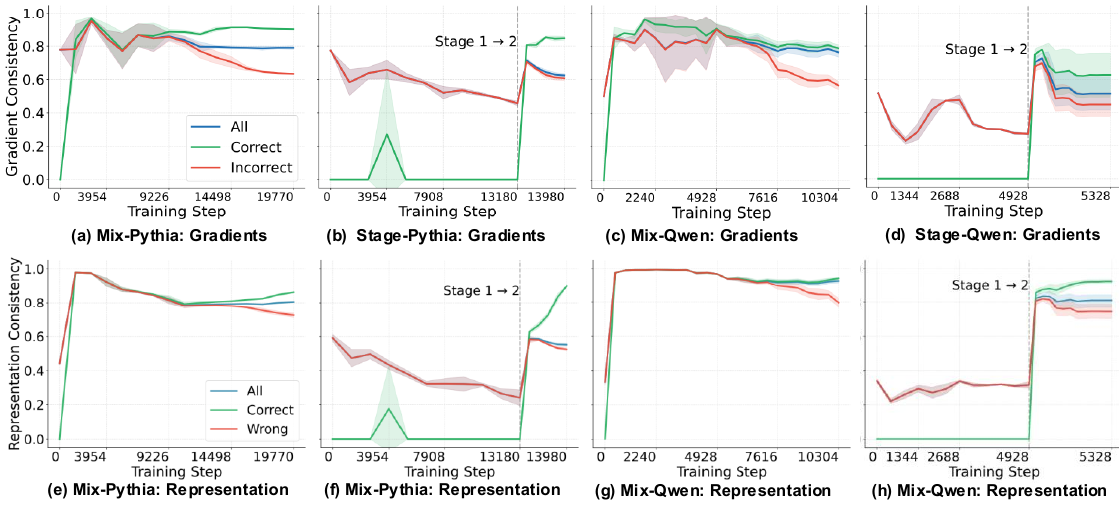}
    \caption{Evolution of gradient and representation consistency for attribute $a_4$ during training Pythia and Qwen.}
    \label{fig:pythia_qwen_training_evolution}
\end{figure*}

\begin{figure*}[h]
    \centering
    \includegraphics[width=0.8\linewidth]{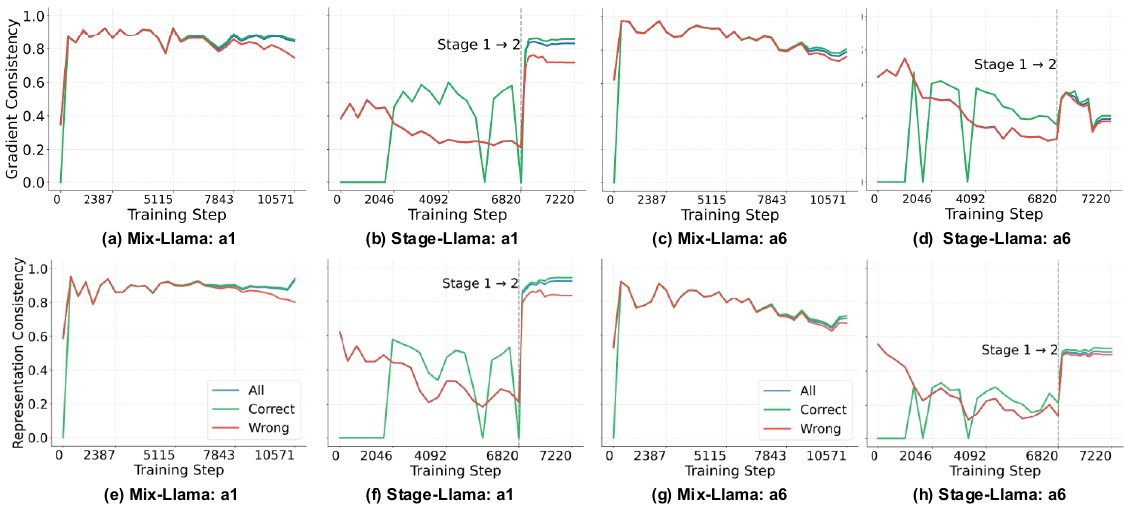}
    \caption{Evolution of gradient and representation consistency for attributes $a_1$ and $a_6$ during training Llama models.}
    \label{fig:llama_training_evolution}
\end{figure*}

\begin{figure*}[h]
    \centering
    \includegraphics[width=1\linewidth]{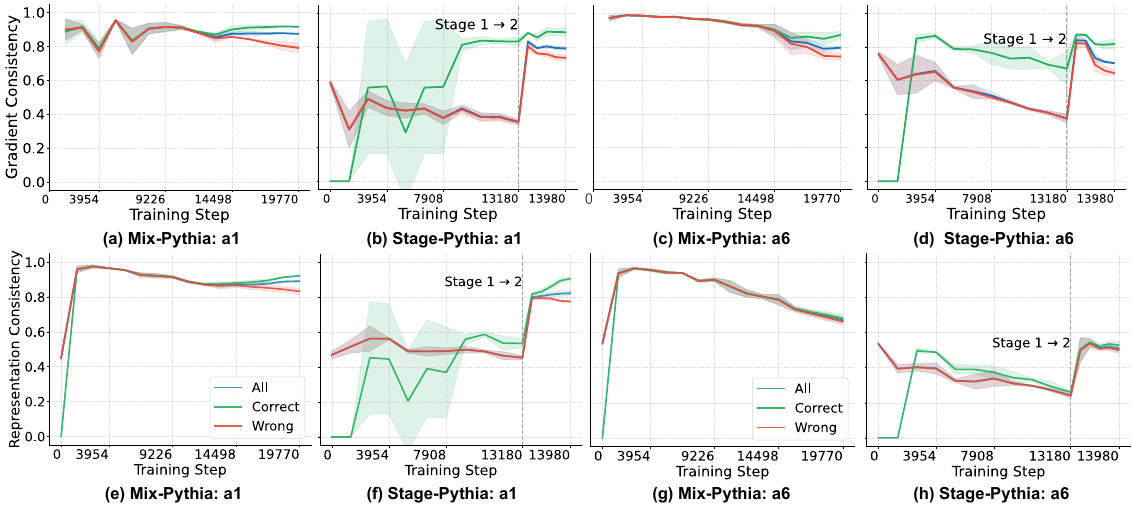}
    \caption{Evolution of gradient and representation consistency for attributes $a_1$ and $a_6$ during training Pythia models.}
    \label{fig:pythia_training_evolution}
\end{figure*}

\begin{figure*}[h]
    \centering
    \includegraphics[width=1\linewidth]{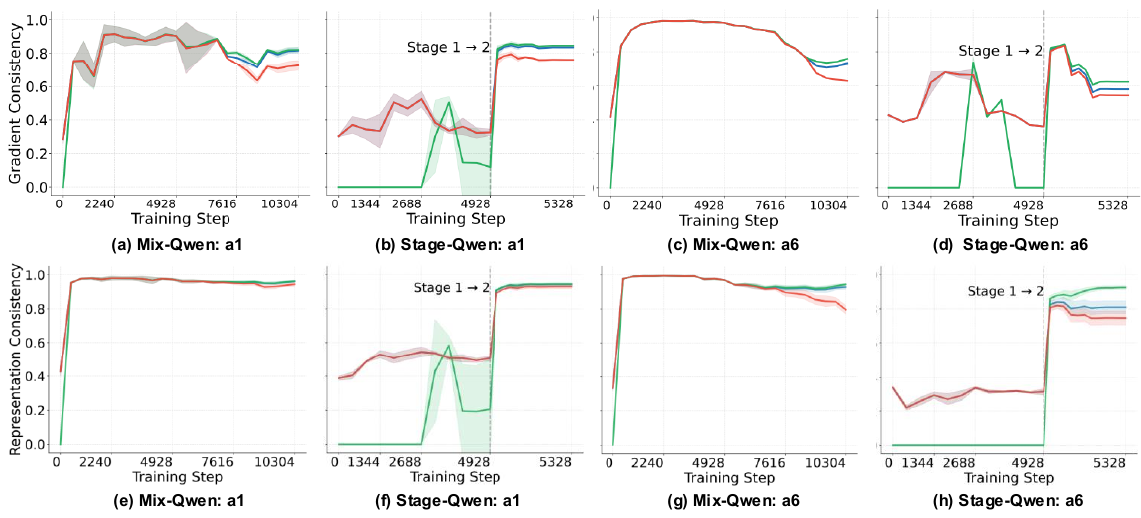}
    \caption{Evolution of gradient and representation consistency for attributes $a_1$ and $a_6$ during training Qwen models.}
    \label{fig:qwen_training_evolution}
\end{figure*}

Figures~\ref{fig:llama_representation_evolution}, \ref{fig:pythia_qwen_training_evolution}, \ref{fig:llama_training_evolution} and \ref{fig:pythia_training_evolution} show the representation and gradient consistency during training.

\clearpage

\subsection{Empirical Analysis of Shared Parameters}
\label{appendix:details_of_shared_parameters}

In this section, we empirically verify that shared parameters emerge as a functional by-product of consistent gradients. We provide a formal definition, the identification procedure, and an ablation study to quantify their impact on fact recall.

\paragraph{Definition and Identification.}
We define shared parameters ($\mathcal{S}$) as top-$k$ model parameters significantly influenced by both the BIO and QA tasks. 
\begin{definition}[Shared Parameters]
Let $\Delta_i^{\text{task}}$ represent the cumulative gradient influence on parameter $\theta_{i}$ during fine-tuning on dataset $\mathcal{D}_{\text{task}}$:
\begin{equation}
    \Delta_i^{\mathrm{task}} = \sum_{d_n \in \mathcal{D}_{\mathrm{task}}} \mathrm{lr}(d_n) \times \mathrm{grad}_{i}(d_n).
\end{equation}
We define $\mathcal{A}_{\text{task}}$ as set of top-$k$ parameters with the highest $\Delta_i^{\text{task}}$.
The \textbf{shared parameters} $\mathcal{S}$ are the intersection of these sets: $\mathcal{S} = \mathcal{A}_{\text{BIO}} \cap \mathcal{A}_{\text{QA}}$.
\end{definition}

To account for data distribution, we further distinguish between:
\begin{itemize}
\item $\mathcal{S}_{\text{\Romannumeral 1}}$ (In-distribution): $\mathcal{A}_{\text{BIO\_in}} \cap \mathcal{A}_{\text{QA\_in}}$.
\item $\mathcal{S}_{\text{\Romannumeral 2}}$ (Out-of-distribution): $\mathcal{A}_{\text{BIO\_out}} \cap \mathcal{A}_{\text{QA\_in}}$.
\end{itemize}

\paragraph{Functional Importance via Grafting.}
To determine if shared parameters are the functional core of fact recall, we compare the model's recall capability with and without the availability of these parameters.
We employ the Grafting approach~\cite{panigrahi2023task} in the following two settings:
\begin{enumerate}
\item \textbf{Baseline Localization}: We identify the minimal set of parameters relevant to fact recall by optimizing a binary mask to recover full model performance. In this setting, the optimizer can select any weights, including shared parameters.
\item \textbf{Ablated Localization}: We first ablate the shared parameters by reverting them to their pre-trained values and freezing them. We then repeat the localization process to find the best possible fact-recall circuit using only the remaining parameters.
\end{enumerate}

The performance difference between these two settings quantifies the specific contribution of shared parameters to fact recall.
A substantial gap would demonstrate that shared parameters and the representation consistency they represent are the primary functional drivers of recall, accounting for the majority of the model's capability.

\paragraph{Experimental Setup.}
Due to computational constraints, $\Delta_i^{\text{task}}$ is estimated using a representative subset of 20 individuals.
We track the top-$10^8$ influenced parameters, excluding embedding and projection layers.
Grafting optimization is performed using SGD for 24 hours per configuration on NVIDIA A100 40GB.
To identify the recall-related circuit, we tune the subset size over $[0, 100\%]$ to find the minimal parameters that recover 92--97\% of the fully fine-tuned model's performance.
For computational efficiency in this analysis, grafting optimization and evaluation are conducted on a subset of 1,000 individuals.

\begin{figure}
    \centering
    \includegraphics[width=1\linewidth]{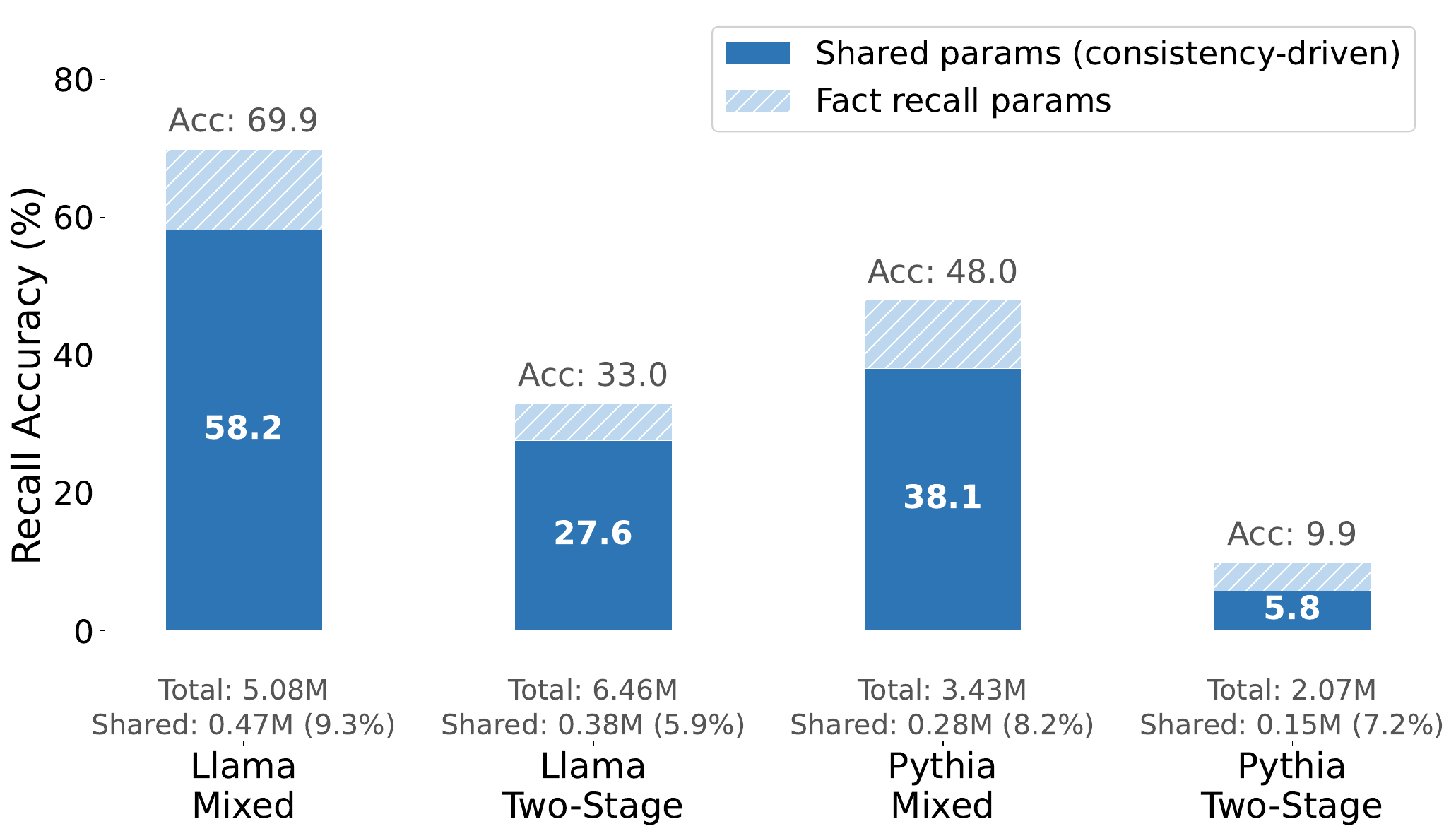}
    \caption{\textbf{Functional impact of shared parameters.} We quantify the contribution of shared parameters as the performance gap between the Baseline (fact recall params) and the Ablated setting.
    The label ``M'' denotes the number of parameters in millions.}
    \label{fig:shared_params_chart}
\end{figure}

\paragraph{Results.}
Figure~\ref{fig:shared_params_chart} illustrates the results.
While shared parameters consistently account for a small fraction of the total recall-related weights (e.g., $\sim$9.3\% of the 5.08M parameters in Mix-tuned Llama), they disproportionately drive over 59\% of the recall accuracy across all tested models.
This demonstrates that shared parameters, as the structural anchor of representation consistency, are the primary functional drivers of successful fact recall.
Furthermore, Mix-tuned models consistently develop a larger pool of shared parameters than their Stage-tuned counterparts.
Crucially, these shared parameters exhibit significantly higher functional efficiency, as evidenced by the higher accuracy-to-parameter ratio (e.g., 58.2\% / 0.47M vs. 27.6\% / 0.38M in the Llama series).
This result aligns with our representation-level observations: mixed training fosters more consistent representations, which in turn manifests as a higher proportion of pivotal shared parameters that drive successful fact recall.

\begin{figure*}[t]
    \centering
    \includegraphics[width=1\linewidth]{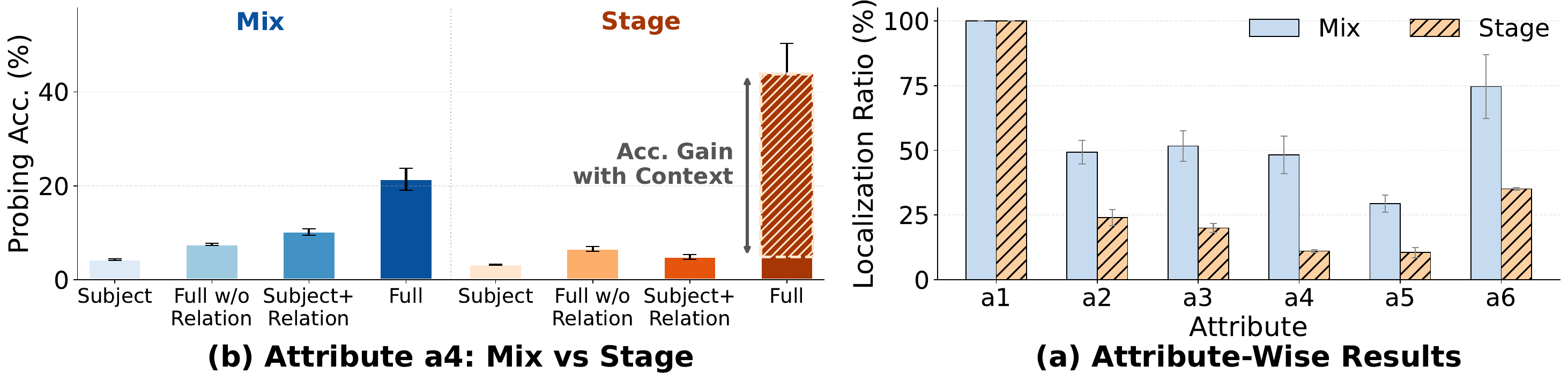}
    \caption{Mechanistic divergence of Pythia models in BIO attribute encoding.}
    \label{fig:pythia_bio_saving_pos_bar}
\end{figure*}

\begin{figure*}[t]
    \centering
    \includegraphics[width=1\linewidth]{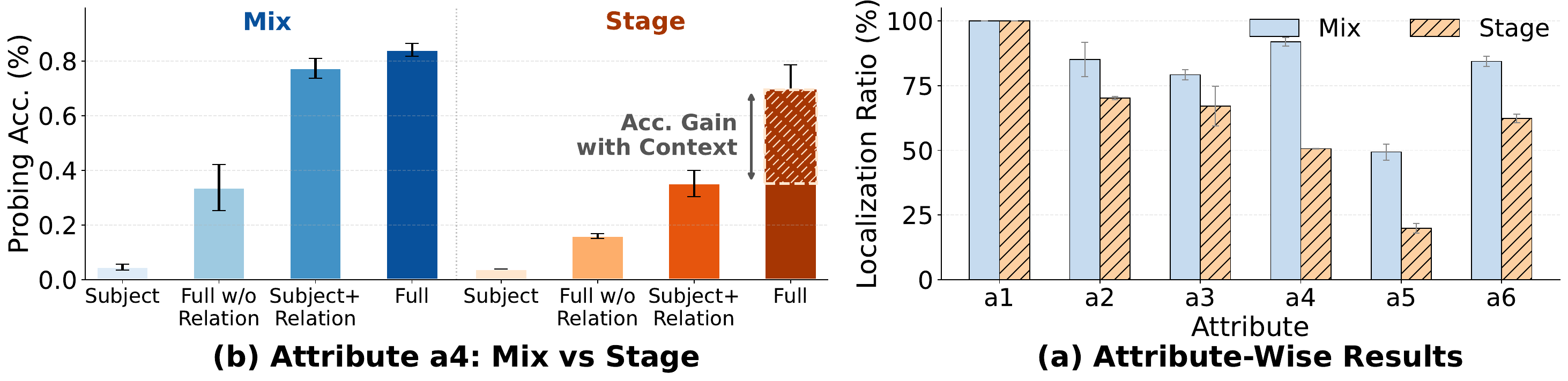}
    \caption{Mechanistic divergence of Qwen models in BIO attribute encoding.}
    \label{fig:qwen_bio_saving_pos_bar}
\end{figure*}

\subsection{Details on BIO Attribute Encoding}
\label{appendix:pythia_bio_attribute_encoding}

To analyze how models encode attributes within BIO-formatted data, we evaluate attribute predictions under four input conditions.

\textbf{Hyperparameters.} We train linear probes on the final hidden states to predict target attributes. We employ the AdamW optimizer with an initial learning rate of 0.001 and a linear decay schedule (no warm-up). The batch size is set to 100. We perform 2,000 training updates for Pythia models and 1,500 updates for Llama models to ensure convergence.
We randomly split 10,000 individuals' BIO into training and evaluation sets with a 9:1 ratio.

Figures~\ref{fig:pythia_bio_saving_pos_bar} and \ref{fig:qwen_bio_saving_pos_bar} presents the performance across these levels for Pythia and Qwen models.
Compared with Two-stage training, mixed training retrieves more attributes when using \textit{Trigger} input, demonstrating its encoding dependence on localized grounding.

\end{document}